\documentclass[10pt,twocolumn,letterpaper]{article}

\usepackage{iccv}
\usepackage{times}
\usepackage{epsfig}
\usepackage{graphicx}
\usepackage{amsmath}
\usepackage{amssymb}
\usepackage{color}
\usepackage[aboveskip=0pt]{subcaption}
\usepackage{placeins}
\usepackage{booktabs} % To thicken table lines

\usepackage[ruled,linesnumbered]{algorithm2e}

\usepackage[pagebackref=true,breaklinks=true,letterpaper=true,colorlinks,bookmarks=false]{hyperref}

\usepackage{array}
\newcommand{\PreserveBackslash}[1]{\let\temp=\\#1\let\\=\temp}
\newcolumntype{C}[1]{>{\PreserveBackslash\centering}p{#1}}
\newcolumntype{R}[1]{>{\PreserveBackslash\raggedleft}p{#1}}
\newcolumntype{L}[1]{>{\PreserveBackslash\raggedright}p{#1}}

\iccvfinalcopy % *** Uncomment this line for the final submission
\definecolor{ben}{rgb}{0.0, 0.58431372549019607843137254901961, 0.65098039215686274509803921568627}

\definecolor{steven}{rgb}{0.03921568627,0.43137254902,0.82352941176}

\definecolor{mat}{rgb}{0.0457464363464789,0.464567523657459760,0.1567890967856745}

\definecolor{greg}{rgb}{0.5457464363464789,0.564567523657459760,0.1567890967856745}

\definecolor{todo}{rgb}{1.0, 0., 0.}

\def\iccvPaperID{18} % *** Enter the ICCV Paper ID here

\ificcvfinal\fi
\begin{document}

%%%%%%%%% TITLE
\title{SteReFo: Efficient Image Refocusing with Stereo Vision}

\author{
\hspace{-3mm}Benjamin Busam$^{\ast,1,2}$\\
\hspace{-3mm}{\tt\small b.busam@tum.de}
\and
\hspace{-3mm}Matthieu Hog$^{\ast,1}$\\
\hspace{-3mm}{\tt\small matthieu.hog@huawei.com}\\
\and
\hspace{-3mm}Steven McDonagh$^{1}$\\
\hspace{-3mm}{\tt\small steven.mcdonagh@huawei.com}
\and
$^{\ast}$equal contribution
\and
\hspace{-3mm}Gregory Slabaugh$^{1}$\\
\hspace{-3mm}{\tt\small gregory.slabaugh@huawei.com}\\
\and
$^{1}$Huawei Noah's Ark Lab\\
\and
$^{2}$Technical University of Munich
}

\maketitle
%\thispagestyle{empty}

%%%%%%%%% ABSTRACT
\begin{abstract}\noindent
%May it be to attract the attention of the viewer on a particular object, give an impression of depth, or simply reproducing the way we perceive scenes as humans, shallow depth of field images are used extensively by professional and amateur photographers.
Whether to attract viewer attention to a particular object, give the impression of depth or simply reproduce human-like scene perception, shallow depth of field images are used extensively by professional and amateur photographers alike. 
%To this end high quality lenses used in DSLR cameras are targeted to focus on specific depth planes by design while producing visually pleasing bokeh.\\
To this end, high quality optical systems are used in DSLR cameras to focus on a specific depth plane while producing visually pleasing bokeh.\\
We propose a physically motivated pipeline to mimic this effect from all-in-focus stereo images, typically retrieved by mobile cameras. It is capable to change the focal plane \emph{a posteriori} at 76~FPS on KITTI~\cite{geiger2012we} images to enable real-time applications.
As our portmanteau suggests, \textit{SteReFo} interrelates stereo-based depth estimation and refocusing efficiently.
In contrast to other approaches, our pipeline is simultaneously fully differentiable, physically motivated, and agnostic to scene content.
It also enables computational video focus tracking for moving objects in addition to refocusing of static images.
We evaluate our approach on publicly available datasets~\cite{geiger2012we,mayer2016large,cordts2016cityscapes} and quantify the quality of architectural changes.

%In this paper we present and compare several learning-based refocusing architectures having the particularity of being fully differentiable, physically-motivated, agnostic to scene content and efficient. Result on real data for the best-working architecture are good in terms of quality and timing.

\end{abstract}

%%%%%%%%% BODY TEXT

\section{Introduction}\noindent
\vspace{-2.0em}
\paragraph{Motivation.}
Around the turn of the millennium, Japanese photographers coined the term \textit{bokeh} for the soft, circular out-of-focus highlights produced by near circular apertures~\cite{prakel2010visual}.
To this day, bokeh is a sign of high quality photographs acquired using professional equipment, closely linked to the depth of field of the optical system in use~\cite{nasse2010depth}.
Historically, producing such photos has been exclusively possible with high-end DSLRs.
Synthesizing the effect of such high-end hardware finds application in particular in consumer mobile devices where the goal is to mimic the physical effects of high-quality lenses in silico~\cite{hauser2018image}.
Due to the inherent narrow aperture of cost-efficient optical systems commonly used in mobile phones, the acquired image is all-in-focus. This property hampers the natural image background defocus often desired in many types of scene capture, such as portrait images.\\
To address this problem a trend has emerged, where shallow depth of field images are computationally synthesized from all-in-focus images~\cite{wadhwa2018synthetic}, usually by leveraging a depth estimation.
In the rest of the paper we refer to this task as \textit{refocusing}.
%The proposed solution is then to reproduce this blur computationally from the narrow aperture, i.e. all in focus, image by leveraging high quality depth estimation.
%The most striking example of this is the portrait mode on recent smartphones.

\begin{figure}[t]
    \centering
    \includegraphics[width=0.48\textwidth]{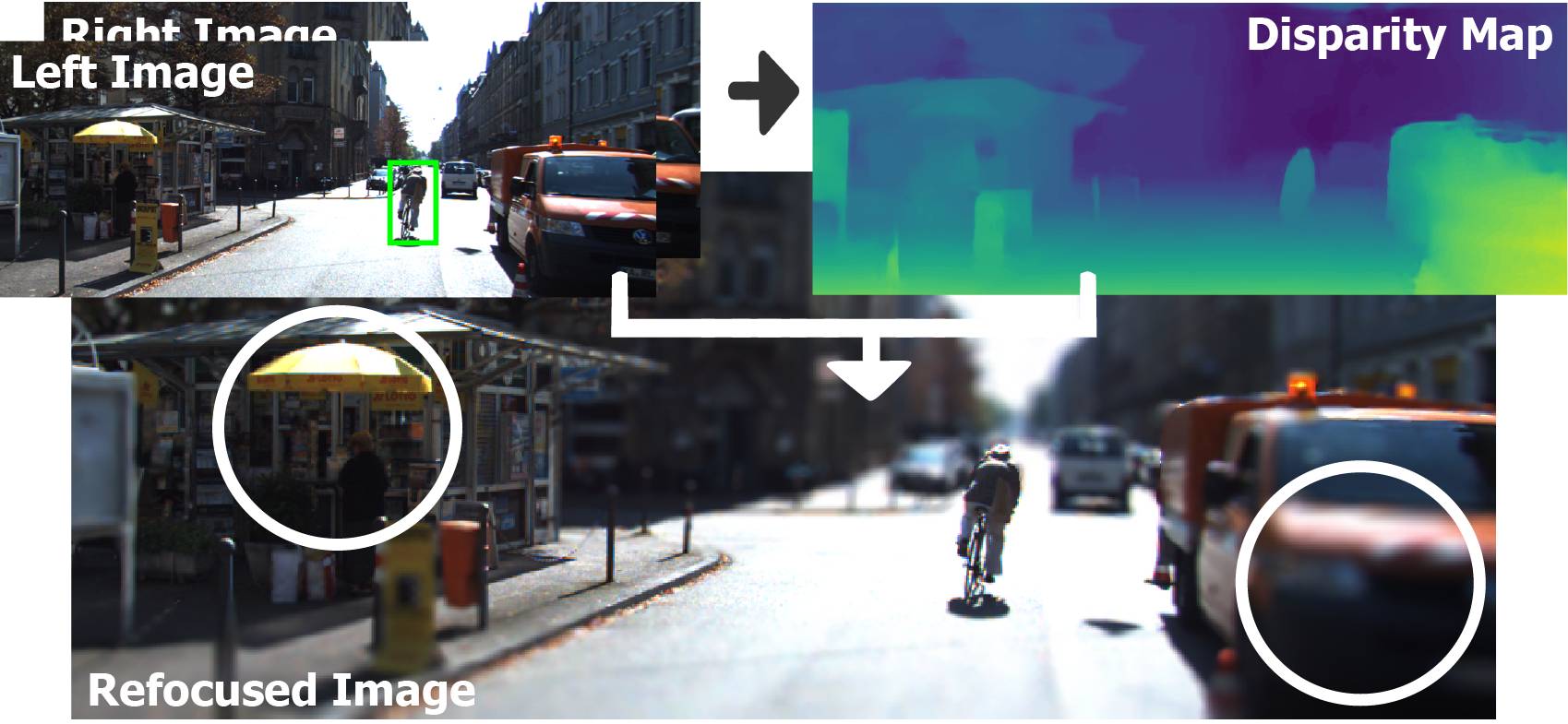}
    \caption{\textit{SteReFo} on video sequence of \cite{geiger2012we}. A disparity map is computed from binocular images while a 2D tracker provides a bounding box (green) to look up the focus depth on the object of interest (the cyclist). With the retrieved depth, the proposed differentiable refocusing stage (white arrow) is utilized to refocus the input frame. The refocused image is in focus in areas that are equal in depth to the cyclist (left) while closer (right) and more distant regions are blurred.}
    \label{fig:fig_one}
\end{figure}

\vspace{-1.0em}
\paragraph{Drawbacks of recent approaches.}
The portrait mode of recent smartphones uses depth estimation from monocular~\cite{wang2018deeplens} or dual-pixel~\cite{wadhwa2018synthetic} cameras. 
To circumvent depth estimation errors, previous approaches rely heavily on segmentation of a single salient object, making them limited to scenes with a unique, predominant region of interest.
%While previous approaches rely heavily on segmentation of a single salient object to compensate for depth estimation errors, they are limited to specific scene types where a single, predominant region of interest is present.
Moreover, this restriction limits the applicability of the underlying refocusing pipeline in other use cases such as object-agnostic image and video refocusing.
%object-agnostic scene refocusing and \mat{content-aware not sure this is clear} \steven{semantic awareness?} video refocusing.

\begin{figure*}[!t]
%\vspace{-1.0em}
\centering
  \includegraphics[width=0.99\textwidth]{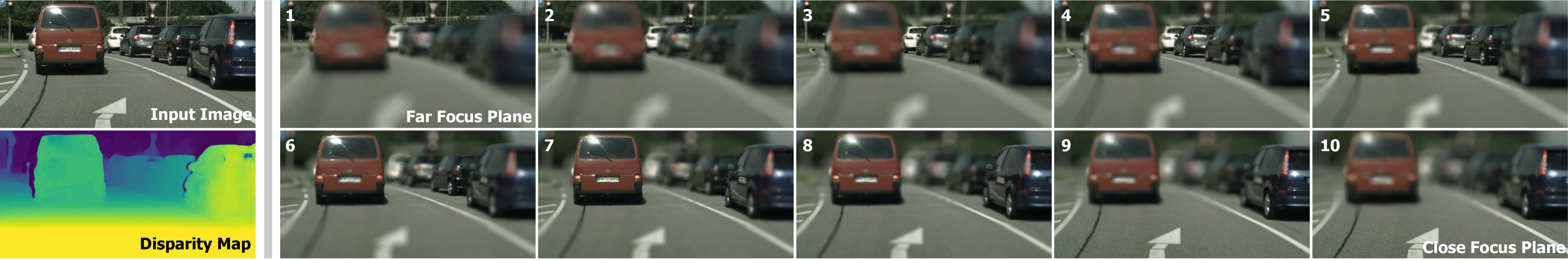}
  \caption{Computational refocusing of an image from \cite{cordts2016cityscapes}. On the left, one input image (from the stereo pair) together with the intermediate disparity map is illustrated while the right part depicts a continuous sequential refocusing on depth planes from far (1) to close (10). Note the smooth transition of the refocus plane, not feasible with segmentation approaches. Also, note the physically motivated radial bokeh effect on the right traffic light in (7), (8) similar to the effects produced by high-end DSLR equipment. Stereo imagery enables in particular high depth precision for sharp boundaries which can be observed e.g. on the right side of the red car.
 }
  \label{fig:teaser}
  \vspace{-1.0em}
\end{figure*}

\begin{figure}[t]
    \centering
    \includegraphics[width=0.48\textwidth]{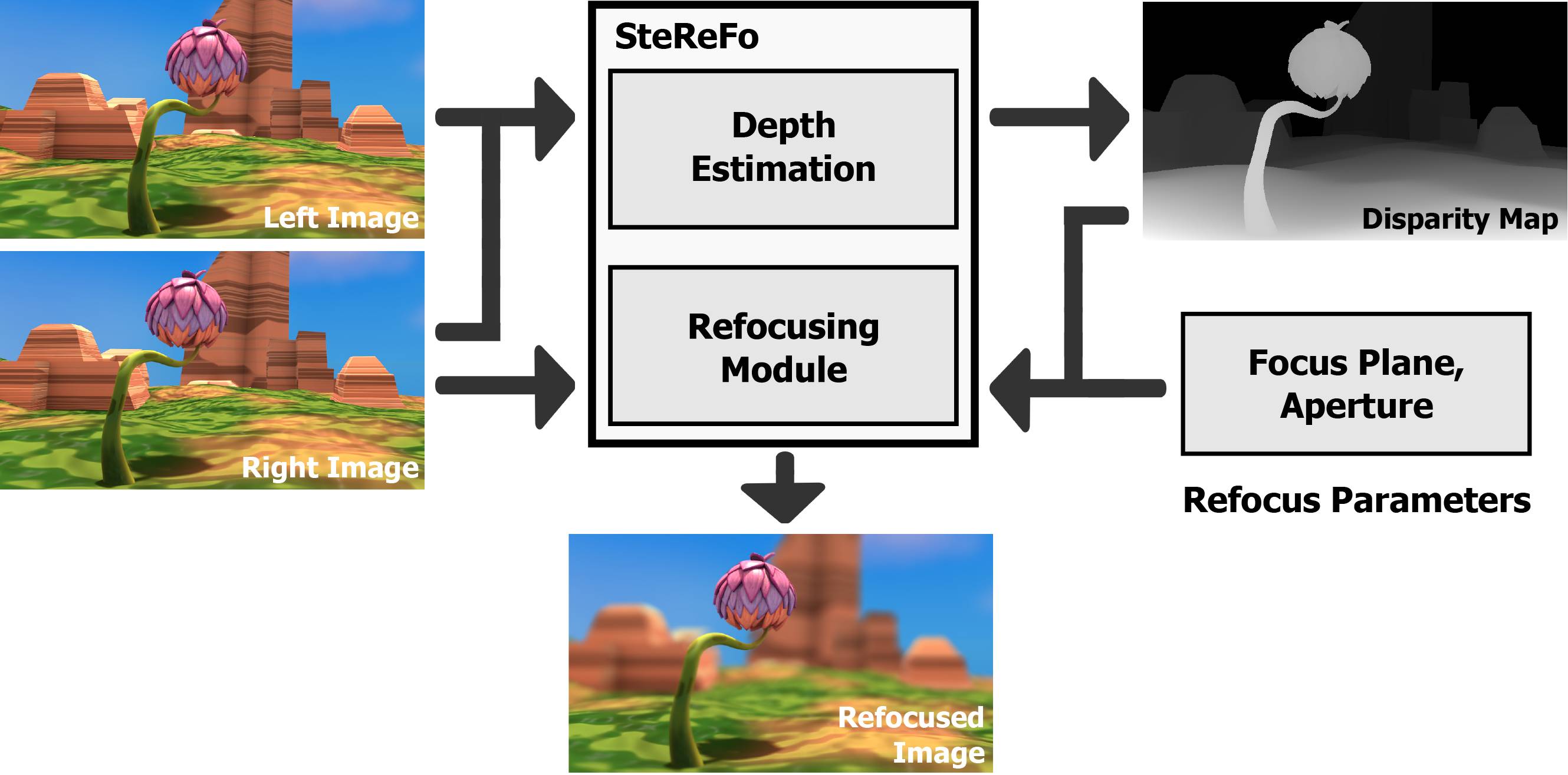}
    \caption{The refocusing pipeline. A stereo pair of all-in-focus images is processed by the depth estimation module which outputs a disparity map. The disparity map together with one input image and the refocus parameters are the input for our efficient refocusing pipeline which leverages the proposed layered depth of field to virtually set a focus plane to refocus the input image.}
    \label{fig:pipeline}
\end{figure}

\vspace{-1.0em}
\paragraph{Contributions and Outline.}
We present a general approach that utilizes stereo vision to refocus images and videos (cf. Fig.~\ref{fig:fig_one}).
Our pipeline, entitled \textit{SteReFo}, leverages the state-of-the-art in efficient stereo depth estimation to obtain a high-quality disparity map and uses a fast, differentiable layered refocusing algorithm to perform the refocusing (Fig.~\ref{fig:pipeline} shows the overall pipeline). A total runtime of $0.14$~sec ($0.11$~sec for depth and $0.03$~sec for refocusing) makes it computationally tractable for portable devices. Moreover, our method is agnostic to objects present in the scene and the user retains full control of both blur intensity and focal plane (cf. Fig.~\ref{fig:teaser}).
We also conduct a study to assess the optimal way to combine depth information with the proposed layered refocusing algorithm. Unlike previous work, we quantify the refocus quality of our methods by means of a perceptual metric.
More specifically, our contributions are:
\vspace{-0.5em}
\begin{enumerate}
    \setlength{\parskip}{0pt}
    \item An efficient pipeline for \textbf{refocusing from stereo images} at interactive frame-rates with a \textbf{differentiable formulation of refocusing} for modular use in neural networks.
    \vspace{-0.2em}
    \item The proposal and study of novel \textbf{architectures to combine stereo vision and refocusing} for physically motivated bokeh.
    \vspace{-0.2em}
    \item Both \textbf{qualitative} and \textbf{quantitative} analysis of our approach on \textbf{synthetic} and \textbf{real} images from SceneFlow~\cite{mayer2016large}, KITTI~\cite{geiger2012we} and CityScapes~\cite{cordts2016cityscapes}.
    \vspace{-0.2em}
    \item A combination of 2D tracking and depth-based refocusing to enable \textbf{computational focus tracking in videos} with tractable computational complexity.
\end{enumerate}
\vspace{-0.5em}
To the best of our knowledge, \textit{SteReFo} is the first method that is jointly trainable for stereo depth and refocusing, made possible by the efficient design of our differentiable refocusing. Our model makes effective refocusing attainable, yet the approach does not require semantic priors and is not limited in blur intensity. We show that it is possible to mimic the manual refocusing effects found in video acquisition systems by autonomous parameter adjustment.

\section{Related Literature}\noindent
Large bodies of work exist in the domains of both vision-based depth estimation and computational refocusing. We briefly review work most relevant to ours, putting our contributions into context.
%To put our work into context, we briefly review the research domains for vision-based depth estimation and computational refocusing.

\subsection{Depth Estimation}\noindent
Depth estimation from imagery is a well studied problem with a long history to perform estimation from image pairs~\cite{scharstein2002taxonomy, lazaros2008review, tippetts2016review}, from temporal image sequences in classical structure from motion (SfM)~\cite{faugeras1988motion, huang2002motion} and simultaneous localization and mapping (SLAM)~\cite{handa2014benchmark, mur2017orb, engel2018direct} and reasoning about overlapping images with varying viewpoint~\cite{agarwal2009building, knapitsch2017tanks}.
In addition, the task of single image depth estimation has shown recent progress using contemporary learning based methods~\cite{laina:2016:deeper, godard:2017:unsupervised, guo:2018:learning, li:2018:megadepth}.
%We round out our depth estimation review by highlighting recent contributions most relevant to the current work.% as an overview and refer the reader to the different works for an additional background.

\vspace{-1.0em}
\paragraph{Monocular vision.}
Deep learning based monocular depth estimation  employ CNNs to directly infer depth from a static monocular image. They are either trained fully supervised (with a synthetic dataset or ground truth from different sensors)~\cite{eigen:2014:depth, laina:2016:deeper} or leverage multiple cameras at training time to use photo-consistency for supervision~\cite{godard:2017:unsupervised, poggi:2018:learning}.
%The first work regarding deep learning based depth estimation comes from Eigen et al.~\cite{eigen:2014:depth}. They employ a CNN to predict a depth map from a static monocular image. Laina et al.~\cite{laina:2016:deeper} improve on this by leveraging a fully convolutional residual network with an efficient encoder-decoder structure.
%These early works are trained in a fully supervised fashion and need to rely on ground truth, either generated synthetically or by means of a different sensing modality.\\
%If multiple cameras are considered during training time, self-supervision with photo-consistency losses becomes possible by warping pixels from one view to the other given a depth estimate. In this way, not only do additional ground truth labels become unnecessary but also the calibration-based error propagation is diminished. Godard et al.~\cite{godard:2017:unsupervised} propose to use left-right stereo consistency for self-supervision while Poggi et al.~\cite{poggi:2018:learning} use trinocular supervision.
%If multiple cameras are considered during training time, self-supervision with photo-consistency losses becomes possible~\cite{godard:2017:unsupervised, poggi:2018:learning}, circumventing the need for ground truth labels and diminishing calibration-based errors. 
However, these approaches are in general tailored for a specific use case and suffer from domain shift errors.%, e.g. if one wants to transfer from indoor data to outdoor scenes or even change the camera design. 
%To address this drawback, Guo et al.~\cite{guo:2018:learning} use stereo matching as a proxy for monocular depth estimation to benefit from pre-training on synthetic data across domains and MegaDepth~\cite{li:2018:megadepth} trains using photos available online and multi-view stereo.
To address this drawback, stereo matching~\cite{guo:2018:learning} or multi-view stereo~\cite{li:2018:megadepth} can be used as a proxy.
While these recent approaches estimate reliable depth values, the depth often suffers from over-smoothing~\cite{godard2018digging} which manifests as ``flying pixels'' in the free space found across depth discontinuities. Accurately and faithfully reproducing such boundaries is, however, critically important for subsequent refocusing quality. Therefore we focus our approach on a binocular stereo cue.

\vspace{-1.0em}
\paragraph{Multi-view prediction.}
For high-accuracy depth maps that preserve precise object boundaries, multiple views are still necessary~\cite{smolyanskiy:2018:importance} and binocular stereo is supported by large synthetic~\cite{mayer2016large} as well as real datasets (cf. KITTI~\cite{geiger2012we}, CityScapes~\cite{cordts2016cityscapes}).
Leveraging this data, StereoNet~\cite{khamis2018stereonet} uses a hierarchical disparity prediction network with a deep visual feature backbone which is capable of running at 60~FPS on a consumer GPU. Its successor~\cite{zhang2018activestereonet} extends the work with self-supervision to the domain of active sensing while maintaining the core efficiency.
We build upon their work to leverage this computational advantage.\\
%Choi et al.~\cite{choi:2018:learning} utilize multiple two-view stereo pipelines to create individual cost volumes with confidence. These can be fused with a depth regression network and Yao et al.~\cite{yao:2018:mvsnet} extend this principle by making use of a differentiable homography warping and a variance-based metric to make the approach agnostic to the amount of input views.\\
More recently, Tonioni et al.~\cite{tonioni2018real} have proposed a way to perform continuous online domain adaptation for disparity estimation with real-time applicability.

\vspace{-1.0em}
\paragraph{Other modalities and applications.}
Fusion of different visual cues can boost accuracy of individual tasks. Leveraging temporal stereo sequences for unsupervised monocular depth and pose estimation, e.g. by warping deep features, improves the accuracy of both tasks~\cite{zhan:2018:unsupervised}. With the same result, Zou et al.~\cite{zou:2018:df} jointly train for optical flow, pose and depth estimation simultaneously while Jiao et al.~\cite{jiao:2018:eccv} mutually improve semantics and depth and GeoNet~\cite{yin:2018:geonet} jointly estimates depth, optical flow and camera pose from video.
Fully unsupervised monocular depth and visual odometry can also be entangled~\cite{zhou:2017:unsupervised} and 3D mapping applications~\cite{zhao:2018:learning} are realized by heavily relying on dense optical flow in 2D and 3D.
Despite the superiority of these approaches, they suffer from larger computational burden or come at the cost of additional training data.

\subsection{Refocusing}
\label{sec:dof_sota}\noindent
%While the end-goal is always the same, \textit{i.e} having depth-dependent blurring of an image, the refocusing diverge in the modality they consider, their quality (physical accuracy, realism, artifact, "niceness") or the speed requirements. 
\textbf{Refocusing algorithms} are heavily utilized in video-games and animated movie production. A plethora of approaches has been proposed for shallow depth of field rendering in the computer graphics community. We follow the taxonomy in \cite{nguyen2007gpu} and refer the reader to ~\cite{barsky2008algorithms} for a complete survey.% of these methods.\\
The first style of approach uses ray tracing to accurately reproduce the ray integration performed in a camera body. While some approaches focus on physical accuracy~\cite{pharr2016physically} and others on (relative) speed~\cite{yang2016virtual}, these methods are very computationally expensive (up to hours per frame %for larger bokeh
).
It is also possible to render a set of views, at 
%slightly
different viewpoints, with fast classical rasterization techniques (\ie creating a light-field \cite{levoy1996light}). Views are then accumulated to produce a refocused  image~\cite{haeberli1990accumulation}. However, this requires the scene to be rendered using an amount of time quadratic in the size of the maximum equivalent blur kernel, which is computationally intractable.\\
This point motivated approaches that seek to reproduce the blur in the image domain directly.
Applying depth-adaptive blur kernels can be formulated as scatter~\cite{krivanek2003fast} or gather~\cite{robison2009image} operations. While the first is hard to parallelize, the latter suffers from sharply blurred object edges and intensity leakage. Moreover, because the blur kernel is different for each pixel, these approaches are hard to optimize for GPUs~\cite{wang2018deeplens}.
Finally, the last class of algorithms represents the scenes as depth layers in order to apply blur with fixed kernels separately~\cite{kraus2007depth}. We give special attention to this type of algorithm in Sec.~\ref{sec:layereddof}.

\begin{figure}[t]
    \centering

    \begin{minipage}{0.45\linewidth}
            \begin{subfigure}[t]{0.87\linewidth}
               \includegraphics[width=\textwidth]{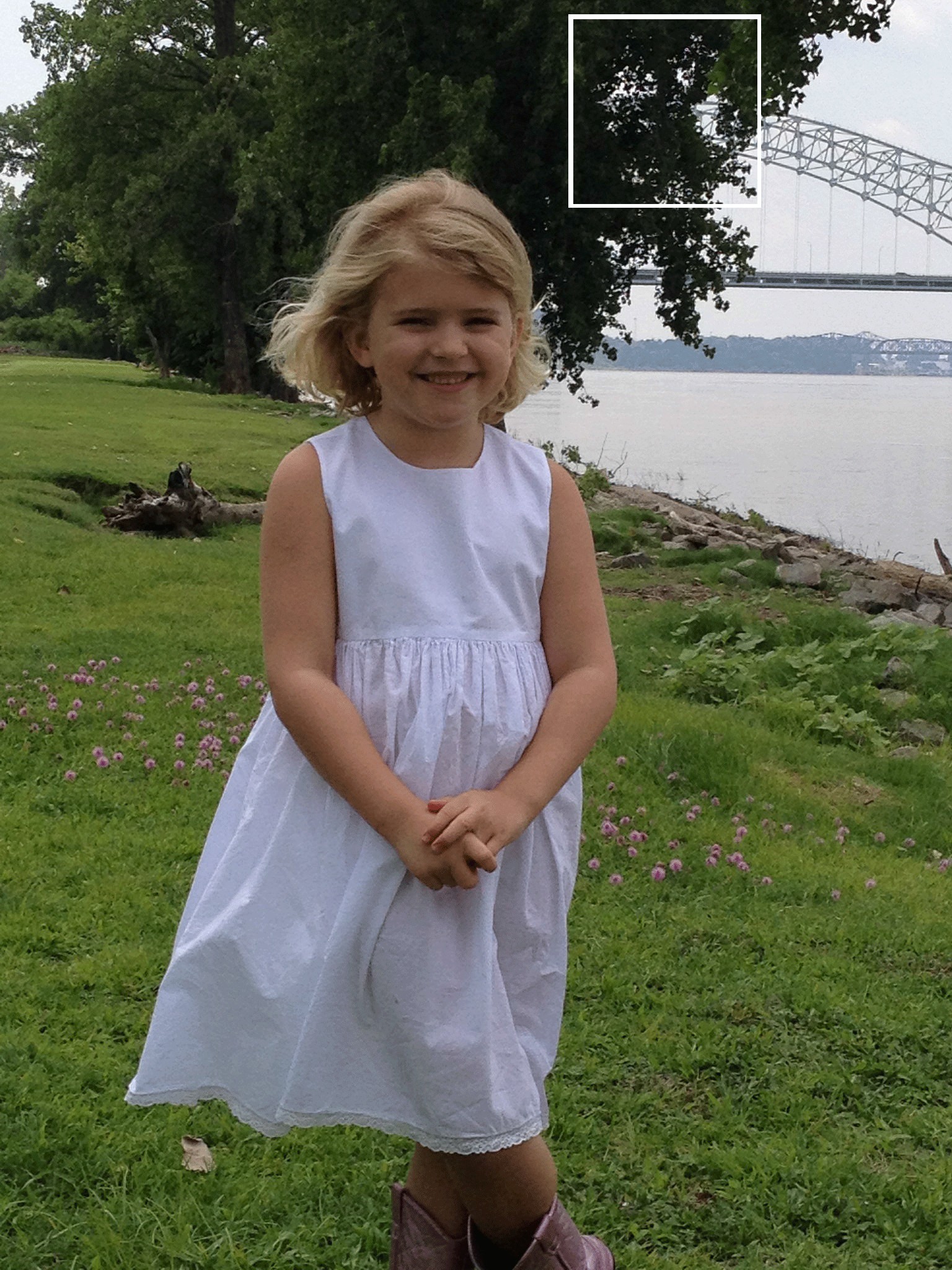}
                \caption{Reference Image}
                \label{fig:deeplens_ref}
            \end{subfigure}
    \end{minipage}
    \begin{minipage}{.52\linewidth}
        \begin{subfigure}{0.45\linewidth}
            \includegraphics[width=\textwidth]{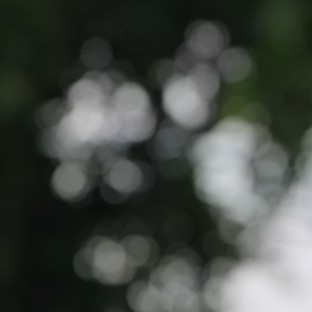}
            \caption{Yang~\cite{yang2016virtual}}
            \label{fig:blur_yang}
        \end{subfigure} 
        \begin{subfigure}{0.45\linewidth}
            \includegraphics[width=\textwidth]{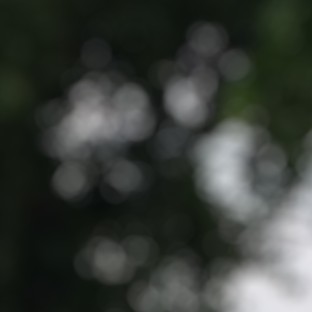}
            \caption{Radial}
            \label{fig:blur_radial}
        \end{subfigure} \\
        \begin{subfigure}{0.45\linewidth}
            \includegraphics[width=\textwidth]{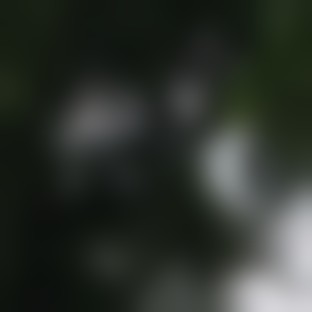}
            \caption{Gaussian}
            \label{fig:blur_gaussian}
        \end{subfigure} 
         \begin{subfigure}{0.45\linewidth}
            \includegraphics[width=\textwidth]{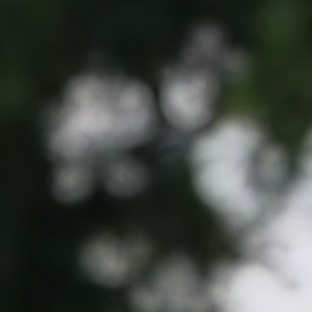}
            \caption{Wang~\cite{wang2018deeplens}}
            \label{fig:blur_wang}
        \end{subfigure} 
    \end{minipage}
    %\vspace{-1em}
    \caption{Refocusing results from different blurring techniques. We display the reference image (\subref{fig:deeplens_ref}) used by Wang et al.~\cite{wang2018deeplens}, and a crop for the different results from a pseudo ray-traced approach \cite{yang2016virtual} (\subref{fig:blur_yang}), a simple radial blur (\subref{fig:blur_radial}), a simple Gaussian blur (\subref{fig:blur_gaussian}) and the result in \cite{wang2018deeplens} (\subref{fig:blur_wang}).
    We observe that the blurred regions for the Gaussian blur and \cite{wang2018deeplens}  lack the distinctive bokeh aspect of DSLR, while physically motivated approaches such as the ray-traced approach and the radial blur, preserve well the bokeh. The latter serves as a backbone in our pipeline (Alg.~\ref{layereddof}).
    }
    \label{fig:bokeh_deeplens}
\end{figure}

\vspace{-1.0em}
\paragraph{Refocusing pipelines.}
In contrast to computer graphics, where scene depth and occluded parts can be retrieved easily, in computer vision, the estimated depth is often noisy and background information is not necessarily available due to the projective nature of cameras.\\
This issue can be addressed through a hole filling task for missing pixel depth~\cite{chou2018occlusion} or by leveraging an efficient bilateral solver for stereo-based depth~\cite{barron2015fast}.% which they compare to a shallow-depth-of-field image generated from a light field. 
 Yu et al.~\cite{yu2011dynamic} directly reconstruct such a light field from stereo images, similar to the techniques discussed previously. It leverages depth estimation, forward warping plus inpainting to reconstruct a reasonable number of views that can be interpolated and summed to render the final image.
The same idea is presented in~\cite{srinivasan2017learning}, but the approach is fully learned with single image input and light field supervision. Zhu et al.~\cite{zhu2017cycle} consider a refocus task using smartphone-to-DSLR image translation. However, authors concede that average performance is considerably worse than highlighted results.\\
More recently, in~\cite{wadhwa2018synthetic} is presented a complete pipeline that computes a person segmentation, a depth map from dual pixel and finally the refocused image. While the results are visually compelling, the method is limited to focusing on a person in the image foreground.
In \cite{wang2018deeplens}, the authors decompose the problem into three modules: monocular depth estimation, blurring and upsampling. While the approach provides visually pleasing images it is unclear how it generalizes, given its fully synthetic training set. Due to the blur step being completely learned, we observed that the images lack the distinctive circular bokeh that professional DSLR cameras produce (see Fig.~\ref{fig:bokeh_deeplens}).\\
Finally, Srinivasan et al.~\cite{srinivasan2018aperture} proposes light field synthesis and supervision with refocused images (i.e. aperture supervision) to learn refocusing. Due to the synthesis of a multitude of views in the first approach, it does not scale well with large kernels. The latter uses the all-in-focus image with a variety of radial kernels and the network is trained to select, for each pixel, which blur value is most likely. The final image is a composition of these blurred images. This approach has the limitation that both the focal plane and the aperture are fixed and cannot be manipulated by a user. In contrast, we want to keep the system parameterizable.
%Secondly, since blur images are computed globally, it is not clear how the approach would perform on images with larger blur radii (esp. in terms of intensity leakage).

\section{Methodology}\noindent
 As illustrated in Fig.~\ref{fig:pipeline}, our pipeline is split into two modules: depth estimation and refocusing. The inputs to the pipeline are a pair of rectified left and right stereo images. A focus plane and an aperture are two user-controllable parameters.
In the following, we explain our proposal for the \textit{SteReFo} module of Fig.~\ref{fig:pipeline} in four different variants which we compare subsequently.

\subsection{Disparity Estimation}\noindent
To produce high quality depth maps, our architecture takes inspiration from two state-of-the-art pipelines for real-time disparity prediction, namely StereoNet~\cite{khamis2018stereonet} and ActiveStereoNet~\cite{zhang2018activestereonet} which estimate a subpixel precise low-resolution disparity map that is consecutively upsampled and refined with RGB-guidance from the reference image.\\
Our depth estimation network consists of two Siamese towers with shared weights that extract deep image features at $1/8$ of the stereo pair resolution following the architecture described in \cite{zhang2018activestereonet}. We construct a cost volume (CV) by concatenation of the displaced features along the epipolar lines of the rectified input images.
The discretization is chosen to include 18 bins.
A shifted version of the differentiable {ArgMin} operator~\cite{kendall2017end} recovers disparities from $i=0$ to $D_{max}=17$ where the disparities are given by
\begin{equation}
    d_{i} = \sum_{d=1}^{D_{max}+1} d \cdot \sigma \left( -C_i \left( d \right) \right) - 1
\label{eq:sec3:disparity}
\end{equation}
with the softmax operator $\sigma$ and the cost $C_i$.
The low resolution disparity map defined in Eq.~\ref{eq:sec3:disparity} is then hierarchically upsampled ($\frac{1}{8} \rightarrow \frac{1}{4} \rightarrow \frac{1}{2} \rightarrow$ full resolution) using bilinear interpolation. Following the idea of Khamis et al.~\cite{khamis2018stereonet}, we use residual refinement to recover high-frequency details. Prior to stacking the resized image and low-resolution disparity map, we pass both individually through a small network with 1 convolution and 3 ResNet~\cite{he2016deep} blocks, as we observed this robustifies our depth prediction quality.\\
The module is trained using a Barron loss~\cite{barron2019more} with parameters $\alpha = 1$, $c = 2$ and RMSProp~\cite{hinton2012neural} optimization with an exponentially decaying learning rate.

\label{sec:layereddof}
\begin{algorithm}[t]
    \SetAlgoLined
    \SetKwInOut{Input}{Input}
    \SetKwInOut{Output}{Output}
    \SetKw{KwBy}{by}
    \Input{All-in-focus Image $I$ with associated disparity map $D$, focus plane $d^f$, aperture $a$, and disparity range $[d_{min},d_{max}]$}
    \Output{Image $I_b$ refocused on the depth plane $d^f$}
     $I_s = [0]$\\
     $M_s = [0]$\\
    \For{$d \gets d_{min}$ \KwTo $d_{max}$ \KwBy $\frac{1}{a}$}
        {
        $M^d=|D-d|<\frac{1}{a}$\\
        $I^d=M^d \circ I$\\
        $r=a \cdot (d-d^f)$\\
        $M^d_b=M^d \ast K(r)$\\
        $I^d_b=I^d \ast K(r)$\\ 
        $M_s = M_s \circ (1-M^d_b)+M^d_b$\\
        $I_s = I_s \circ (1-M^d_b)+I^d_b$
        }
 $I_b = I_s \oslash M_s$
 \caption{The layered depth of field base algorithm used in our approach.}
 \label{layereddof}
\end{algorithm}

\subsection{Efficient Layered Depth of Field}\noindent
Our refocusing module utilizes layered depth of field rendering to enable efficient refocusing. The core idea of layered depth of field rendering~\cite{kraus2007depth} is to first decompose the scene into depth layers in order to separately blur each layer before compositing them back together.
In contrast to ~\cite{wang2018deeplens}, which learns kernel weights, this physically motivated choice directly reflects the effects obtained by DSLR lenses while providing an appropriate balance between efficiency and accuracy for our runtime requirements.
Using this approach, the blur operation is applied by combining fixed-kernel convolutions, that make it very efficient in practice due to contemporary GPU convolutional implementations.\\
% thanks to the progress in implementing GPU convolution triggered by the deep learning era~\cite{chetlur2014cudnn}
We describe the algorithm in Alg.~\ref{layereddof}, where the $\circ$ and $\oslash$ notation are used for the entrywise Hadamard product and division respectively, and $\ast$ denotes convolution. We start from an all-in-focus image $I$ with its associated disparity map $D$, a user-set focus plane $d^f$, an aperture $a$ and a disparity range $[d_{min},d_{max}]$ defined by the stereo setup capabilities.  $I_s$ and $M_s$ are two accumulation buffers. We sweep the scene from back to front within a given disparity range using a step size of (optimally) $\frac{1}{a}$. 
%This step can be reduced to lower the computational burden at the expense of blur area discontinuities.
A mask $M^d$, defining the zones within a disparity window around the disparity plane $d$, is used to extract the corresponding texture of objects within depth plane $I^d$.
The corresponding blur radius $r$ is computed from the distance of the focal plane $d^f$ to the current depth plane $d$ and the given aperture $a$.
The extracted mask and texture are blurred with a radial kernel $K(r)$ of diameter $r$.
The blurred mask and texture are accumulated in the buffers $I_s$ and $M_s$, overwriting the previous values where the mask is not 0, in order to handle clipping in the blur (i.e. prevent out of focus regions to bleed into in-focus regions). The final blurred image $I_b$ is rendered by normalizing the accumulated blur texture with the accumulated masks. 

\vspace{-1.0em}
\paragraph{Adaptive downsampling.}
We alter the base algorithm described in Section~\ref{sec:layereddof} in two ways. To further increase the efficiency of the pipeline, we set a maximum kernel size $k_{max}$ that may be applied. 
For a given disparity plane $d$, we resize the input image by a factor of $\gamma = \left \lceil{2r+1}\right \rceil / k_{max}$ and apply the convolution with a kernel size $K(\gamma \cdot r)$. 
The blur result is then upsampled to full resolution using bilinear interpolation. While this is an approximation, the visual difference is marginal due to its application to out of focus regions. However, the computational efficiency is improved by several orders of magnitude (cf. Sec.~\ref{sec:real_exp}).  

\vspace{-1.0em}
\paragraph{Differentiablility.}
The second modification we carry out is making this algorithm differentiable in order to use it in an end-to-end trainable pipeline. 
In Alg.~\ref{layereddof}, it can be observed that all operations carried out are differentiable, except for computation of the mask which relies on the non-smooth \textit{less than} operator in line 4.
By expressing this operator using the Heaviside step function, the mask computation can be written as: 
\begin{equation}
    M^d  = H\left( \frac{1}{a} - \left| D-d \right| \right) \text{ where } H(x)=
    \begin{cases} 
      0 & \text{x$<$0}\\
      1 & \text{x$\geq$0}
    \end{cases}
\label{eq:heavyside}    
\end{equation}
While the Heaviside step function itself is non-differentiable, a smooth approximation is given by $\hat H(x)=\frac{1}{2}+\frac{1}{2} \tanh(x)$.
Hence we can replace line 4 in Alg. \ref{layereddof} with  
\begin{equation}
    \hat M^d  = \frac{1}{2} + \frac{1}{2} \tanh \left( \alpha \cdot \left( \frac{1}{a} - \left| D-d \right| \right) \right)
\end{equation}
where $\alpha$ controls the transition \textit{sharpness} of the Heaviside step function approximation. We empirically set $\alpha = 10^3$. 

\subsection{Refocusing Architectures}
\label{sec:archis}
\begin{figure*}[t]
    \centering
    \includegraphics[width=\textwidth]{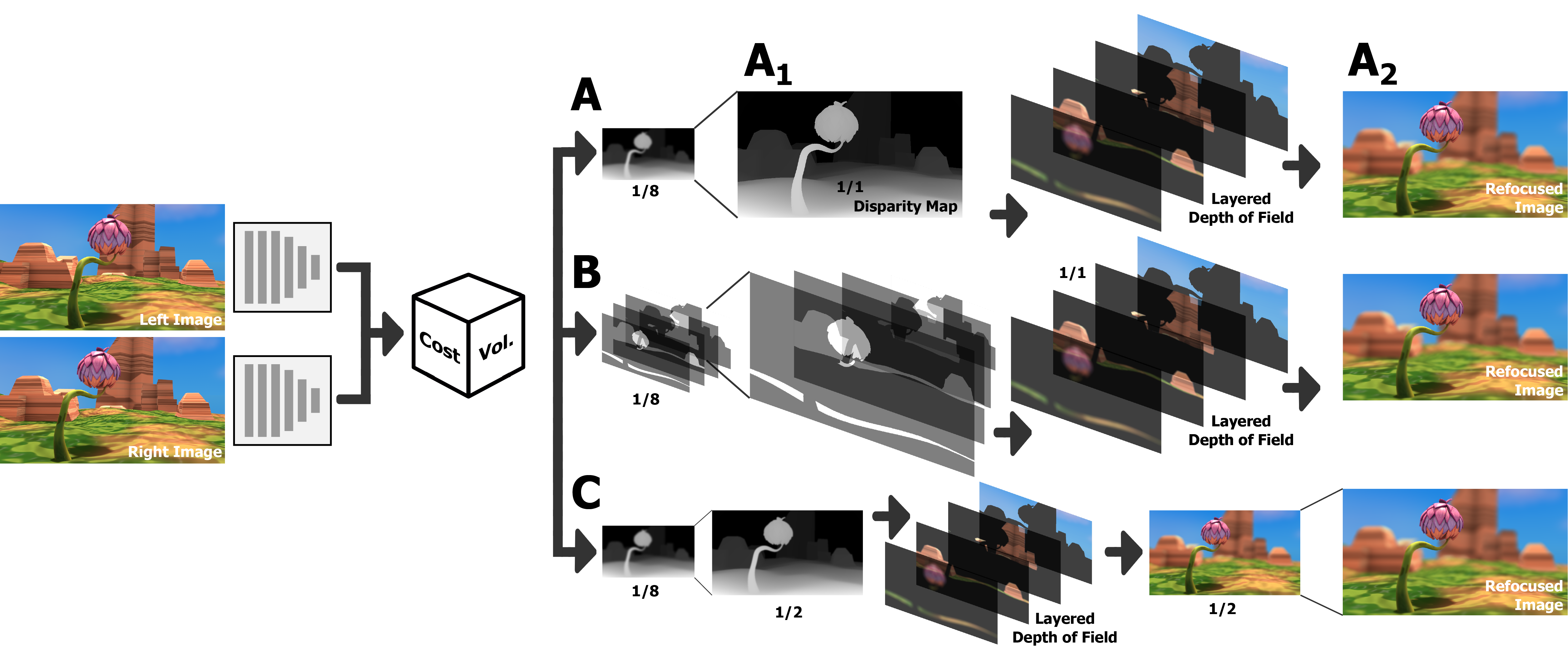}
    \vspace{-1.0em}
    \caption{\textit{SteReFo} architectures. A Siamese tower extracts deep features from a stereo pair which form a cost volume. Four instantiations of our pipeline are depicted. Branch \textbf{A}: the cost volume is then sequentially processed by a depth estimation and differentiable refocusing module and trained with disparity ($A_1$) or aperture ($A_2$) supervision. Branch \textbf{B}: cost volume refinement. Cost volume slices are adaptively upsampled and fed into the layered depth of field pipeline. Branch \textbf{C}: a low resolution depth map is used to predict a downsized refocus image which is consecutively upsampled.}
    \label{fig:architectures}
\end{figure*}

\noindent
To intertwine stereo depth estimation and refocusing in \textit{SteReFo}, we investigate the four architectures illustrated in Fig.~\ref{fig:architectures}.
The first ($A_1$), dubbed \emph{Sequential depth}, takes the disparity, estimated from the stereo network, at full resolution and uses it in the layered depth of field technique described in Section~\ref{sec:layereddof}. While the first part is supervised with the ground truth depth, the second stage is not learned.\\
\emph{Sequential aperture} ($A_2$) is a variation of the first architecture where aperture supervision is used to train the network end-to-end from the blur module. A ground truth blurred image is used instead of the depth, and the loss is defined in the final image domain by applying a pixel-wise Euclidean loss. This is possible thanks to the differentiability of the refocusing algorithm. We use an image refocused with the ground truth disparity for supervision.\\
The third technique, $B$, leverages the fact that the cost volume of the stereo network provides a scene representation very similar to the layer decomposition used in the blurring algorithm. We use each slice of the cost volume (after the StereoNet ArgMin step) directly as a mask $M^d$. We note these slices are of low resolution (1/8 of the input) and therefore bilinearly upsample and refine them using a network with shared weights, up to the resolution required to apply the blurring convolution with a kernel of maximum size $k_{max}$. 
To train this network, we again use images blurred with the ground truth depth map and supervise with an $L_2$ loss. A pretrained StereoNet is utilized for which the weights are frozen before the cost volume computation step.
The refinement network uses the same blocks as in a refinement scale of ActiveStereoNet~\cite{zhang2018activestereonet}. We call this method \emph{cost volume refinement}.\\
Branch $C$ depicts a \emph{blur refinement} for which we propose to start from the $\frac{1}{2}$ resolution depth map provided by a Stereo\-Net intermediate step, blur the image at half resolution, and then upsample the blurred images back to full resolution using an upsampling akin to~\cite{zhang2018activestereonet}. Once again, the network is trained with an $L_2$ loss on a ground truth blurred image and the weights of the StereoNet part, up to the second refinement scale are frozen.

\section{Experimental Evaluation}\noindent
In the following section we provide qualitative and quantitative analysis of our approach on synthetic and real imagery using the public datasets SceneFlow~\cite{mayer2016large}, KITTI~\cite{geiger2012we} and CityScapes~\cite{cordts2016cityscapes}.
All experiments are conducted on an \textit{Intel(R) Core(TM) i7-8700} CPU machine at 3.20~GHz and we trained all neural networks until convergence using an \textit{NVidia GeForce GTX 1080 Ti} GPU with Tensorflow~\cite{abadi2016tensorflow}.

\paragraph{Qualitative Evaluation Metrics.}
Comparing the quality of blurred images is a very challenging task. Barron et al.~\cite{barron2015fast} propose to utilize structural metrics to quantify image quality with a light field ground truth. This modality is difficult to acquire and is therefore usually not present in real datasets of trainable size. Classical image quality metrics, like PSNR and SSIM, do not fully frame the perceptual quality of refocused images~\cite{blau20182018}.
Because there is no consensus on what, quantitatively, makes for a good refocused images (bokeh-wise but also in terms of object boundaries and physical blur accuracy), subjective assessments are often used~\cite{hauser2018image} and some papers exclusively focus on qualitative assessments~\cite{yu2011dynamic,yang2016virtual,srinivasan2017learning,wang2018deeplens,srinivasan2018aperture}.
%Nonetheless, we recognize that the absence of a proper metric inhibits objective comparison of refocusing algorithms even more present due to the lack of a common refocusing dataset. While this clearly defines a research topic in itself, the design and evaluation of a suitable metric is considered beyond the scope of the current work.
In order to provide quantitative evaluation of our results, in addition to classical metrics, we propose to utilize a perceptual metric commonly used by the super resolution community~\cite{blau2018perception, blau20182018}, the NIQE score~\cite{mittal2013making}.\\
In a first experiment, we use a synthetic dataset to train and test the four different approaches described in Section~\ref{sec:archis}. In a second experiment we assess how our pipeline performs on real data.

\begin{figure*}[t]
    \centering
    \begin{tabular}{ccccc}
    Ground Truth & Sequential Depth ($A_1$) & Sequential Aperture ($A_2$) & CV Refinement ($B$) & Blur Refinement ($C$)\\
 
    \includegraphics[width=0.2\textwidth]{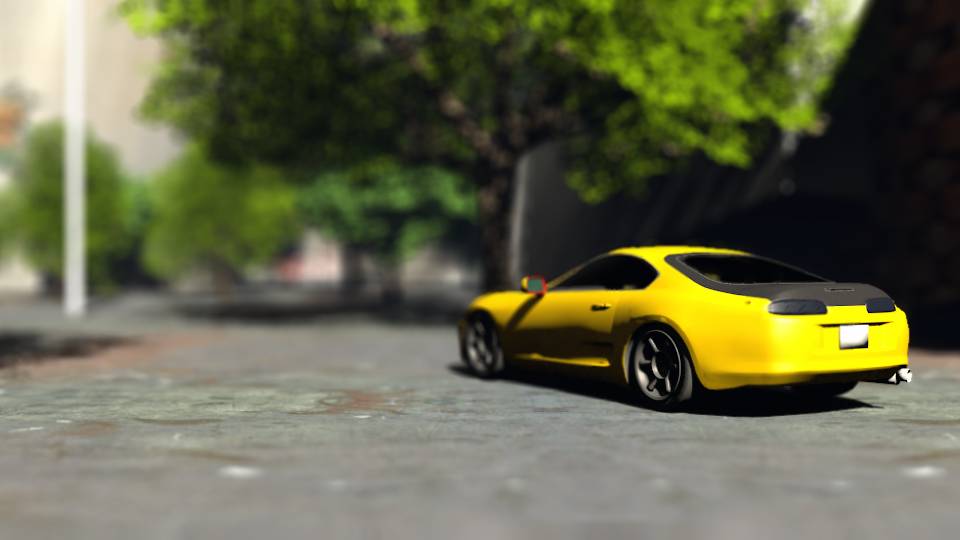}&\includegraphics[width=0.2\textwidth]{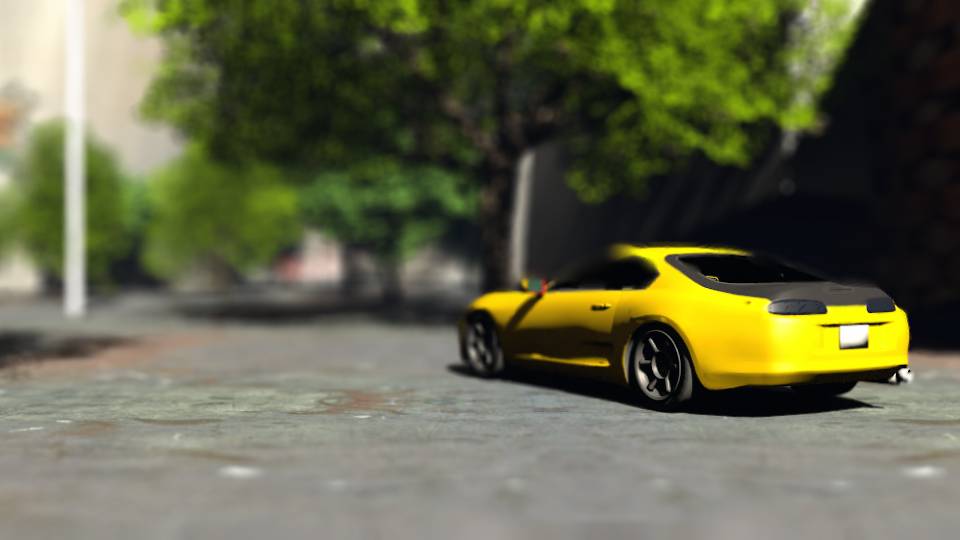}&\includegraphics[width=0.2\textwidth]{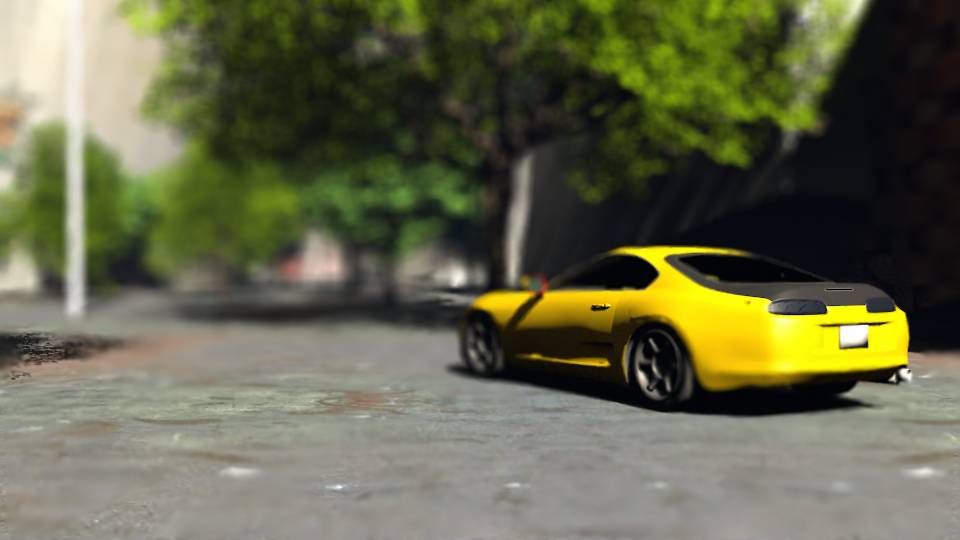}&\includegraphics[width=0.2\textwidth]{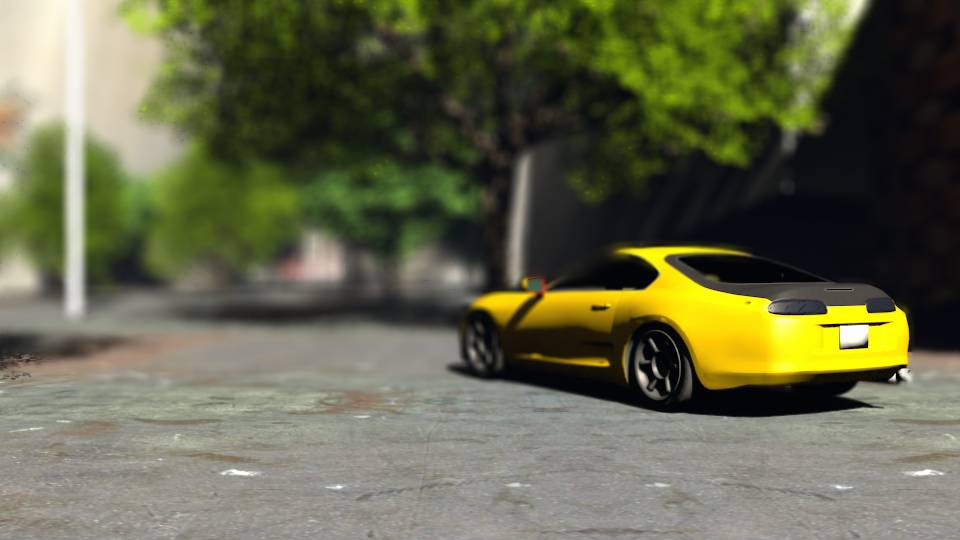}&\includegraphics[width=0.2\textwidth]{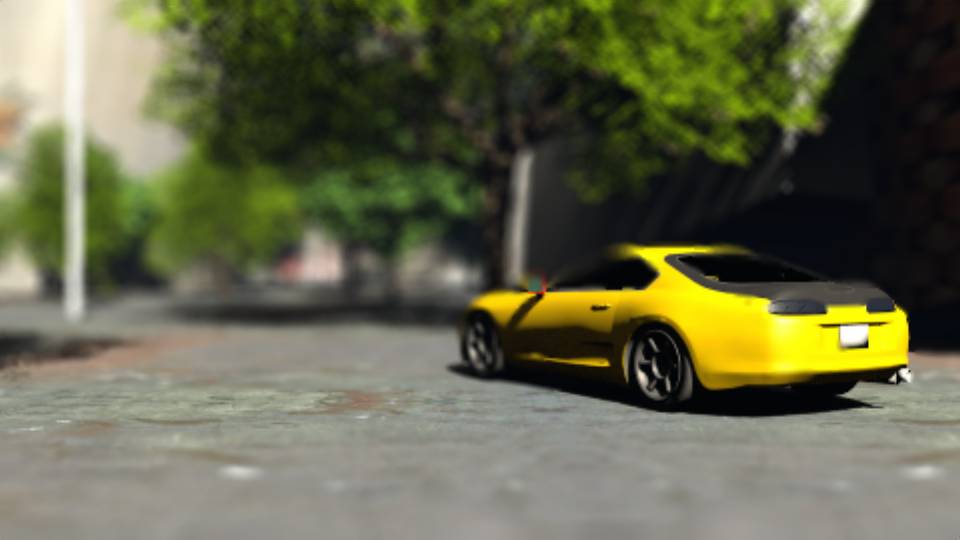}\\

    \includegraphics[width=0.2\textwidth]{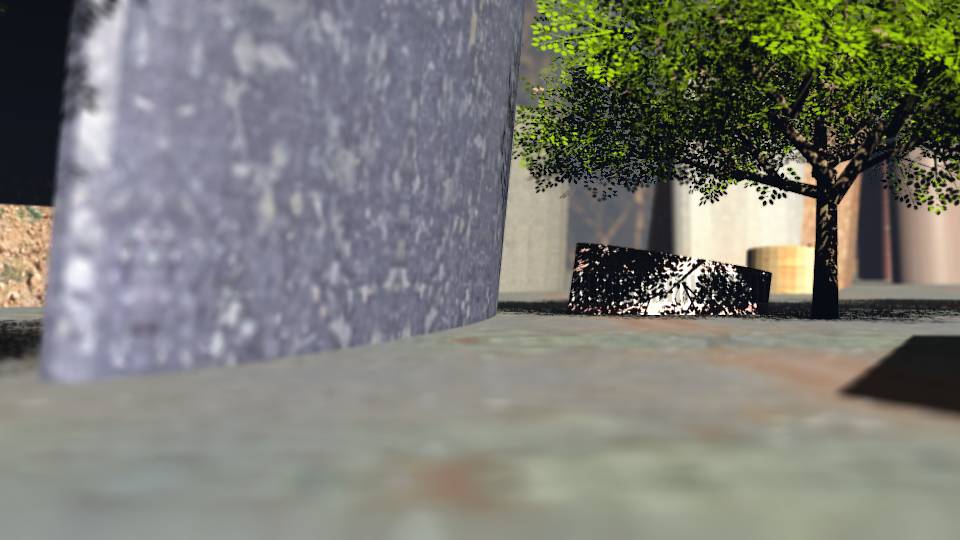}&\includegraphics[width=0.2\textwidth]{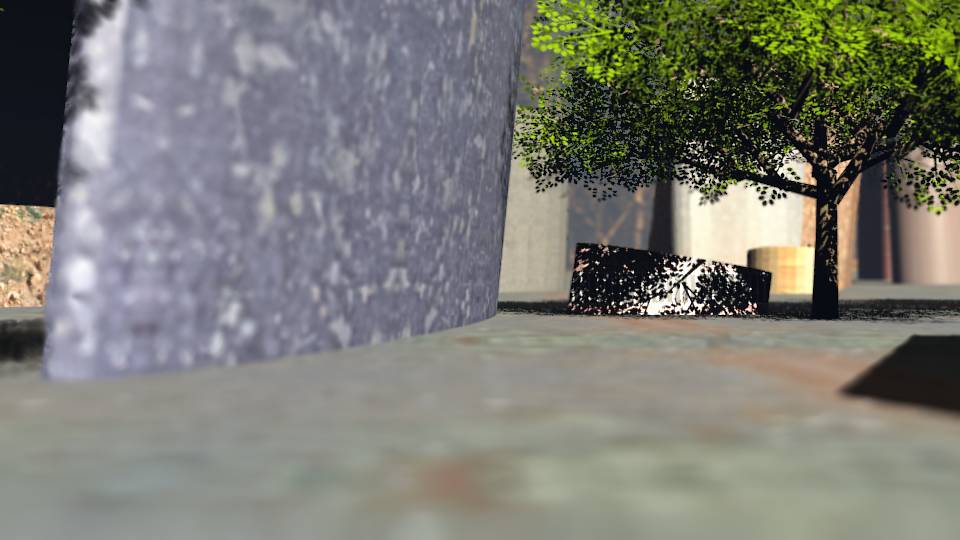}&\includegraphics[width=0.2\textwidth]{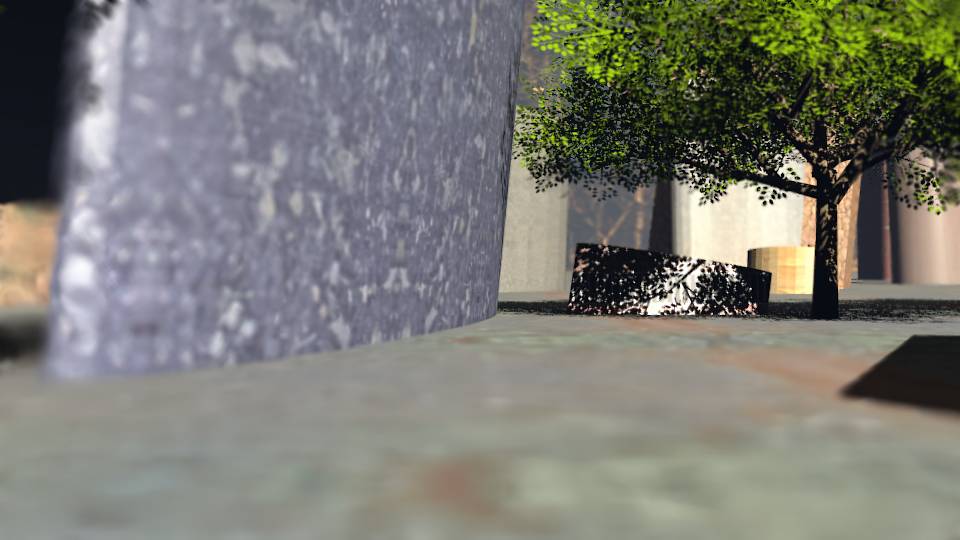}&\includegraphics[width=0.2\textwidth]{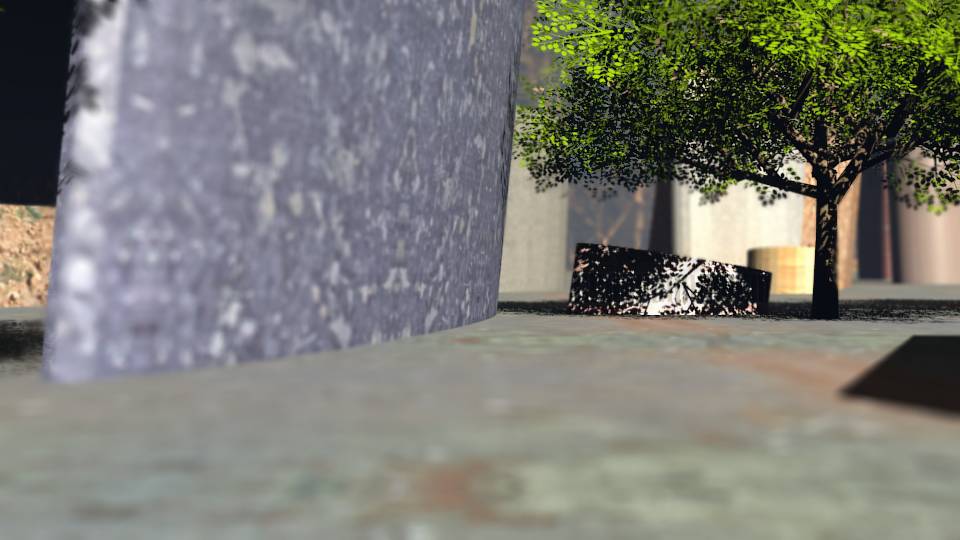}&\includegraphics[width=0.2\textwidth]{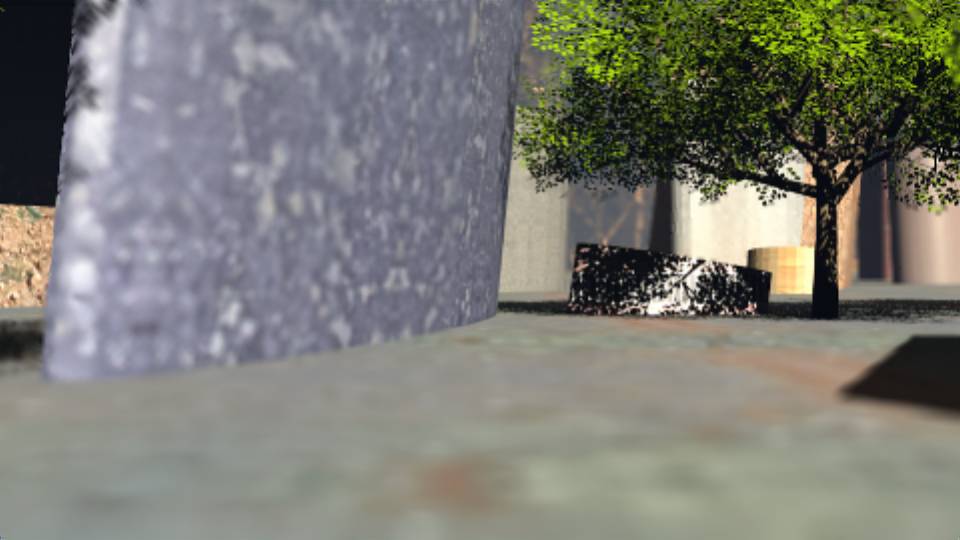}\\

    \includegraphics[width=0.2\textwidth]{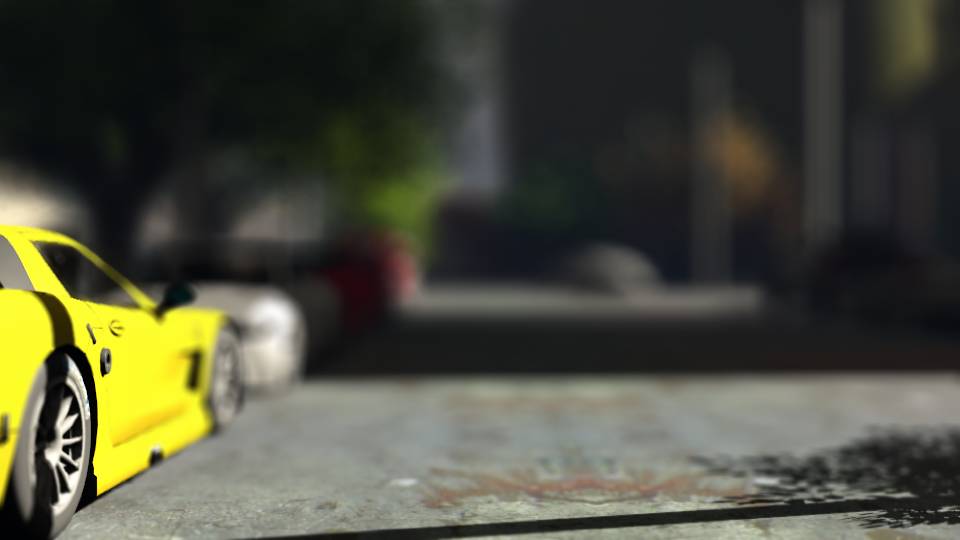}&\includegraphics[width=0.2\textwidth]{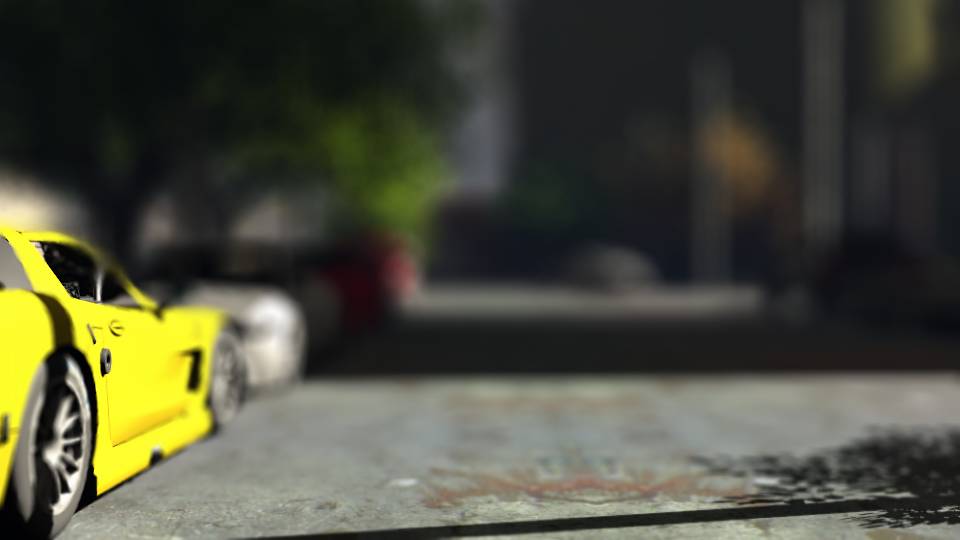}&\includegraphics[width=0.2\textwidth]{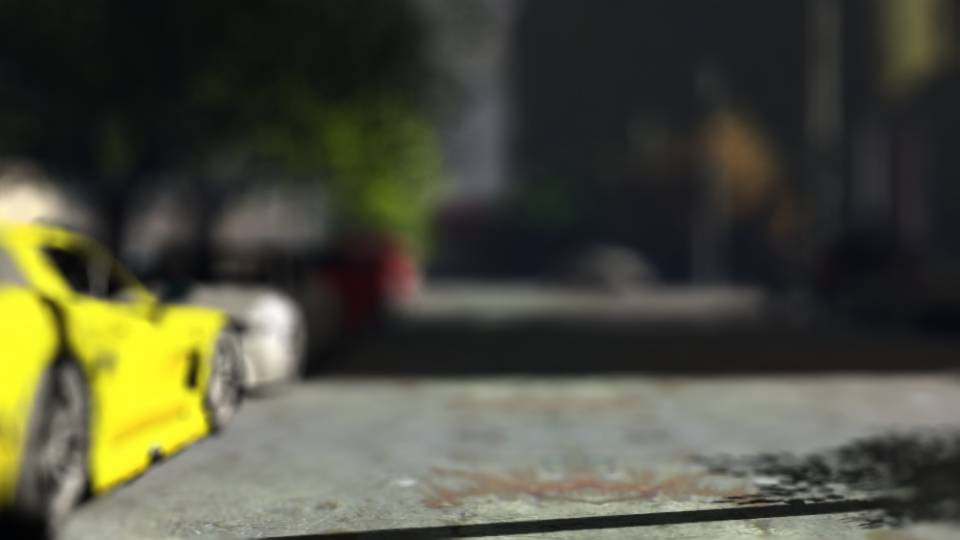}&\includegraphics[width=0.2\textwidth]{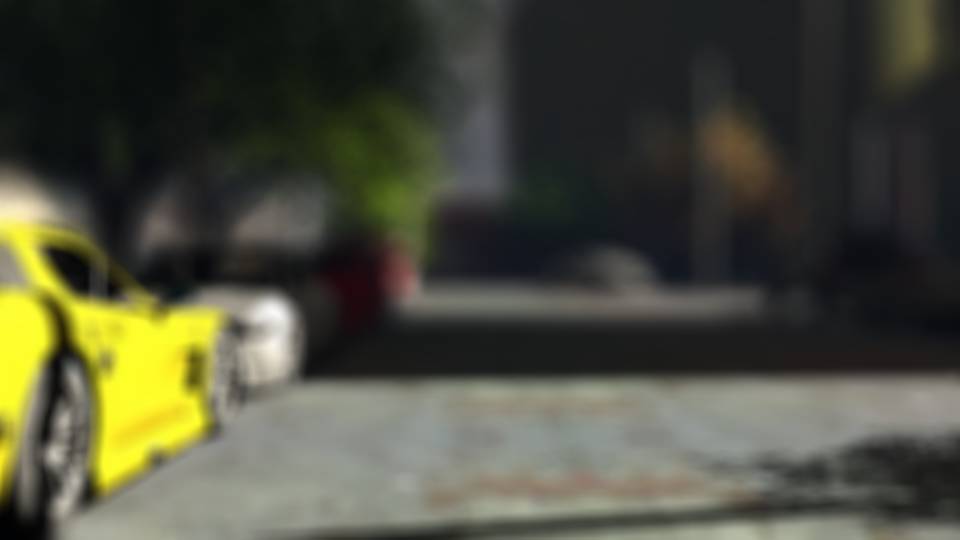}&\includegraphics[width=0.2\textwidth]{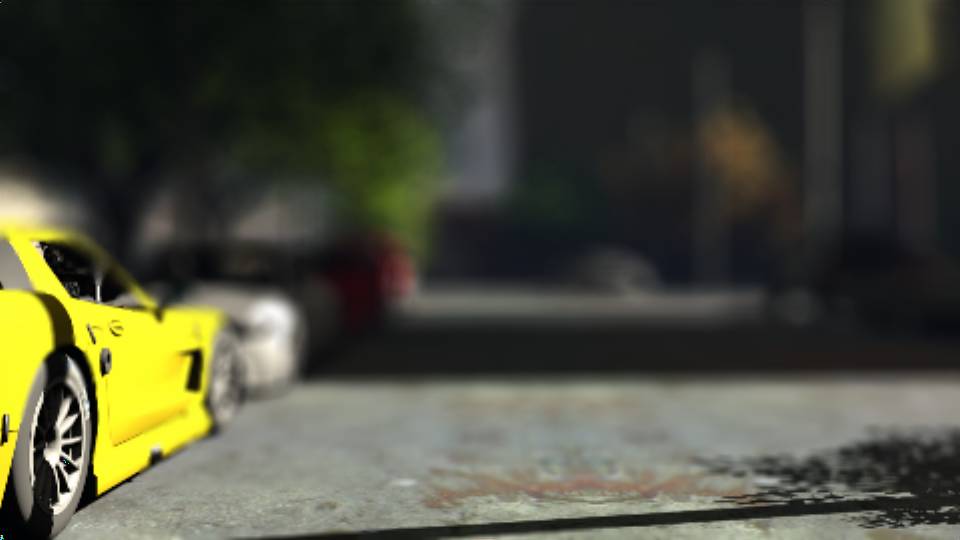}\\

    \end{tabular}
    %\vspace{-1.0em}
    \caption{Comparison of the tested approaches. Each column correspond to one method and each row to one test image. We display on the very left column a ground truth image refocused using the provided ground truth disparity. While we invite the reader to zoom-in to see the details, more example images are included in the supplementary material.}
    \label{tab:comparison_img}
\end{figure*}

\subsection{Architecture Comparison on Synthetic Data}
\begin{figure}[t]
    \centering

    \begin{subfigure}[t]{0.19\linewidth}
        \includegraphics[width=\textwidth]{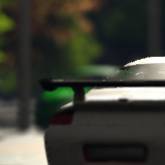}
        \caption{G.T.}
        \label{fig:crop_gt}
    \end{subfigure}
    \begin{subfigure}[t]{0.19\linewidth}
        \includegraphics[width=\textwidth]{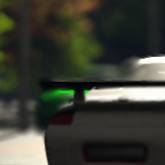}
        \caption{Seq. D.}
        \label{fig:crop_seq}
    \end{subfigure}
    \begin{subfigure}[t]{0.19\linewidth}
        \includegraphics[width=\textwidth]{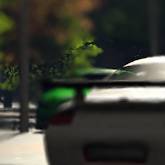}
        \caption{Seq. A.}
        \label{fig:crop_seqap}
    \end{subfigure}
    \begin{subfigure}[t]{0.19\linewidth}    
        \includegraphics[width=\textwidth]{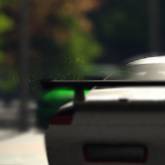}
        \caption{CV Ref.}
        \label{fig:crop_cv}
    \end{subfigure}
    \begin{subfigure}[t]{0.19\linewidth}    
        \includegraphics[width=\textwidth]{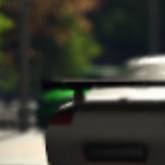}
        \caption{Blur Up.}
        \label{fig:crop_up}
    \end{subfigure}
    %\vspace{-1.0em}
    \caption{Crop on a representative artifact for the proposed methods. (\subref{fig:crop_gt}) is the ground truth, (\subref{fig:crop_seq}) the output of the sequential approach with depth supervision, (\subref{fig:crop_seqap}) the sequential approach trained with aperture supervision, (\subref{fig:crop_cv}) refocusing from the cost volume, (\subref{fig:crop_up}) the blur upsampling.}
    \label{fig:crop_comparison}
\end{figure}

\begin{figure}[t]

\centering
\begin{tabular}{ccc}
Ground Truth & Depth Supervis. & Aperture Superv.\\

\includegraphics[width=0.15\textwidth]{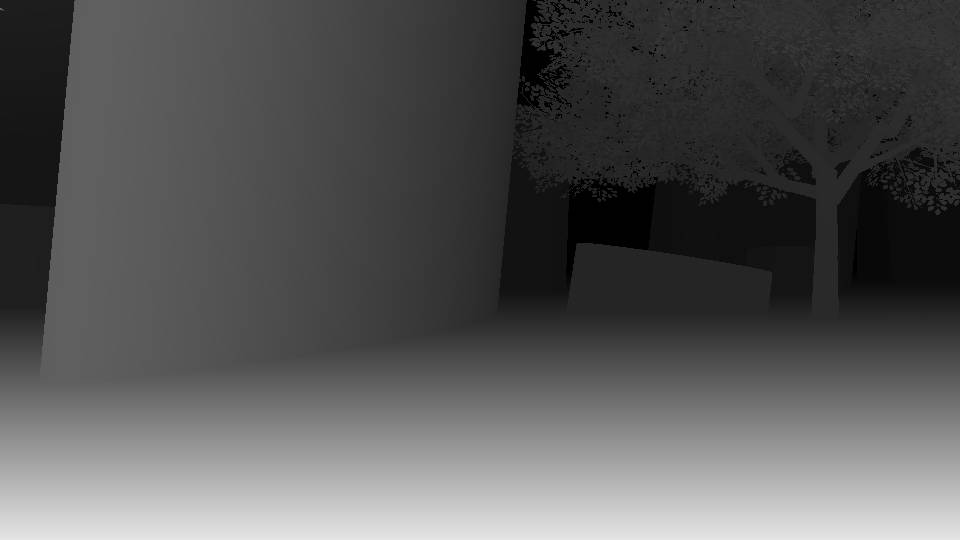}&\includegraphics[width=0.15\textwidth]{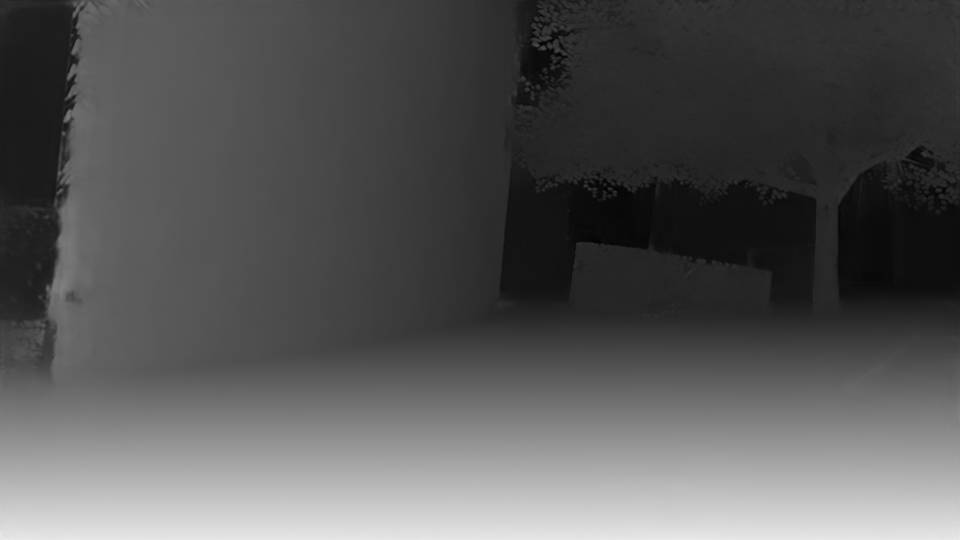}&\includegraphics[width=0.15\textwidth]{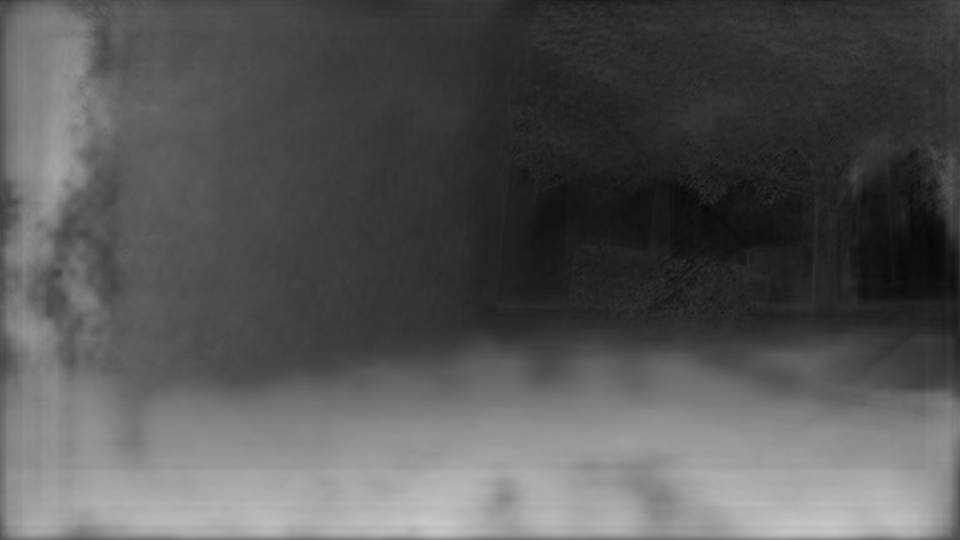}\\

\end{tabular}
%\vspace{-1.0em}
\caption{Depth map comparison. Left to right columns correspond to ground truth disparity, the disparity from depth supervision and from aperture supervision, respectively.}
\label{tab:comparison_depth}
\end{figure}

\noindent
We train the introduced approaches on the full \textit{35mm driving} set of SceneFlow~\cite{mayer2016large} and exclude $11$ frames for testing.
The virtual aperture and focal plane is fixed to $a = 0.1$ and $d^f = 100$, while the disparity range is set to $d \in [0, 300]$. The maximum blur kernel size is $k_{max} = 11$.\\
In Fig. \ref{tab:comparison_img} we display the result of the forward pass on our test images and display a representative crop example in Fig.~\ref{fig:crop_comparison}.\\
Qualitatively we notice that, overall, the result of the sequential approach $A_1$ outperforms the other three in terms of boundaries, bokeh appearance, and blur accuracy. The blur upsampling method $C$ produces blurry output, even in areas that are intended to be sharp, and we observed a loss in the bokeh \textit{circularity}. The cost volume refinement approach $B$, although a conceptually interesting idea, was found to introduce some high frequency artifacts in the blurred zones and also has generally lower quality boundaries. Finally, aperture supervision ($A_2$) is by far the worst of the approaches, qualitatively, as we find high sensitivity to uniform areas in the image in addition to poor performance at object boundaries.\\
We further investigate the source of the quality drop in Fig.~\ref{tab:comparison_depth}, where we compare the output of the depth using ground truth depth supervision and ground truth blurred images (i.e. aperture supervision). While the depth supervision retrieves disparity precision in particular along depth discontinuities, the supervision with aperture fails to recover small details and depth boundaries, ultimately destroying the depth map gradients.

\vspace{-1.0em}
\paragraph{Quantifying the result.}
The NIQE score~\cite{mittal2013making} unifies a collection of statistical measures to judge the visual appearance of an image.
We initially evaluate our introduced approaches for the test images numerically in Tab.~\ref{table:niqe} and analyze the absolute difference from the ground truth retrievals for a relative measure.\\
On inspection of this result, we observe that our sequential supervision with depth provides the best quantitative performance on all considered metrics which is in line with recent findings~\cite{godard2018digging} that show artifact removal for simple depth estimation models.
%The relative standard deviation for the perceptual metrics is one order of magnitude below the other architectures and this observation is consistent with our qualitative analysis.
Approaches $B$ and $C$ are on par while the blur refinement $C$ was found to have the highest (worst) relative score of $2.4$ distance from the ground truth with a better structural similarity.
While the NIQE score aids discovery of best performing methods for this problem ($A_1$ vs. others), it is not well correlated with our visual judgment of the aperture supervision result. We believe this is due to the fact that the aperture supervision image is indeed wrongly refocused, however, it does not show many high-frequency artifacts in contrast to the blur refinement which is also reflected in the classical metrics SSIM and PSNR.

\begin{table}[t]
\centering
    %\scalebox{1.0}{
    \begin{tabular}{lR{1.65cm}R{1.85cm}R{1.65cm}R{1.7cm}}
        & Sequential & Seq. Apert. & CV Ref. & Blur Ref.\\ \toprule
        %NIQE & \bf{6.47$\pm$1.44} & 7.31$\pm$2.35 & 8.19$\pm$3.42 & 8.14$\pm$1.75\\
        NIQE~$\downarrow$ & \bf{6.5$\pm$1.4} & 7.3$\pm$2.4 & 8.2$\pm$3.4 & 8.1$\pm$1.8\\
        %$\sigma$ & \bf{1.44} & 2.35 & 3.42 & 1.75\\
        %Diff. & \bf{0.12$\pm$0.04} & 1.16$\pm$0.82 & 1.72$\pm$1.05 & 2.38$\pm$1.84\\
        Rel.~$\downarrow$ & \bf{0.1$\pm$0.04} & 1.2$\pm$0.8 & 1.7$\pm$1.1 & 2.4$\pm$1.8\\
        \midrule
        SSIM~$\uparrow$ & \bf{0.98$\pm$0.01} & 0.95$\pm$0.02 & 0.95$\pm$0.03 & 0.96$\pm$0.01\\
        PSNR~$\uparrow$ & \bf{39.16$\pm$1.1} & 36.25$\pm$1.2 & 36.60$\pm$1.8 & 36.56$\pm$1.5\\
        \bottomrule
    \end{tabular}%}
    %\vspace{-1.0em}
    \caption{Evaluation results for perceptual~\cite{mittal2013making} and structural metrics. $\downarrow$ indicates that lower, $\uparrow$ that higher is better.}
    \label{table:niqe}
%\vspace{-1.0em}
\end{table}

\vspace{-1.0em}
\paragraph{Discussion.}
We believe our experimental work gives valuable insight into how the tasks of depth and refocusing can be entangled and the resulting benefits of doing so.\\
Firstly, it suggest that depth supervision, and therefore high quality depth data, is essential for refocusing, even more so than retrieving images that are numerically close to the ground truth. This is quantitatively supported by the retrieved NIQE scores.
Secondly, upsampling and refining the depth gives better quantitative and qualitative results than upsampling the blurred image. This suggests that the task of correcting the depth is superior to adjusting a blurred image with residual refinement, especially in the boundaries of in-focus objects.\\
Counter-intuitively, upsampling and filtering the cost volume reveals to be a difficult task, and while the results are still visually appealing, the high computational complexity makes this approach less tractable.

\subsection{Results on Real Data}
\label{sec:real_exp}
\begin{figure}[t]
    \centering
    \includegraphics[width=\linewidth]{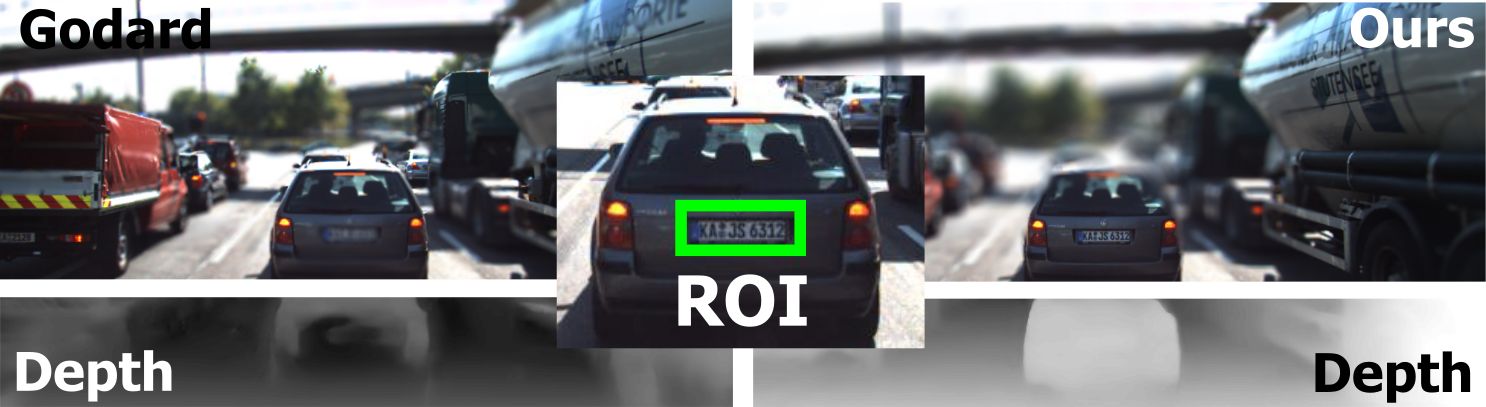}
    %\vspace{-1.0em}
    \caption{Experiments on real data. The region of interest is set with 2D tracking onto the number plate (middle) and the corresponding depth value is recovered. The left illustrates refocusing using a depth map generated by \cite{godard:2017:unsupervised} while we show on the right the result of our sequential refocusing pipeline, together with the underlying disparity map.
    }
    \label{fig:kitti}
\end{figure}

\noindent
The lack of a publicly available dataset for stereo-based refocusing approaches and the requirements of recent methods for additional information such as segmentation masks~\cite{wang2018deeplens}, varying aperture~\cite{srinivasan2018aperture} or co-modalities given by dual-pixel sensing~\cite{wadhwa2018synthetic} and light fields~\cite{barron2015fast} impede a standardized evaluation protocol. In order to assess how our approach performs on real data, we utilize that our pipeline does not require these additional cues and use the datasets proposed in~\cite{cordts2016cityscapes} and \cite{geiger2012we}. We pick our best-performing approach, i.e. the sequential pipeline using depth supervision ($A_1$) and pretrain on CityScapes~\cite{cordts2016cityscapes} to perform static image refocusing (cf. Fig.~\ref{fig:teaser}) and refine with the coarse KITTI~\cite{geiger2012we} ground truth for temporal evaluation.\\
To examine the efficiency of our pipeline, we apply SteReFo individually on consecutive frames of \cite{geiger2012we}. We utilize correlation-filter based 2D object tracking~\cite{valmadre:2017:2dtracker} to reason about the spatial location of objects of interest, per frame, prior to applying image refocusing.
The 2D tracker provides state-of-the-art performance at high frame rates, a choice that makes a lightweight overall pipeline feasible.
%2D tracker formulation makes use of the Fourier domain and provides state-of-the-art performance at high frame rates. Choice of tracker makes potential future pipeline deployment to lightweight architecture feasible.
For the sake of comparison, we also retrained the monocular depth estimation approach in~\cite{godard:2017:unsupervised}, and refocus the video using the generated depth as input for our layered refocusing pipeline.\\
For both approaches, we use disparity values of $d \in [0, 80]$ and an aperture of $a = 0.25$, the focal plane is defined as the median value of the disparities inside the tracking bounding box, and $k_{max} = 11$ (cf. Fig.~\ref{fig:fig_one}).
We compare the results directly in Fig.~\ref{fig:kitti} and notice that blurring is significantly less consistent with respect to the scene geometry in the case of \cite{godard:2017:unsupervised} compared to our approach. Indeed, the background is defocused as if it was not at infinity, the cars appear to reside in the same depth plane and the lower section of the car in the middle of the image is blurred (where it should not).
The disparity maps for each approach support these observations.
The interested reader is referred to the full video sequence \url{https://youtu.be/sX8N702uIag}.

\vspace{-1.0em}
\paragraph{Timing.}\noindent
We evaluate the timing of our approach on real data.
%The running average running time for the entire pipeline is  $0.142608s$, including $0.113028s$ to compute the disparity and $0.029580s$ for refocusing.
The average runtime for the entire pipeline is  $0.14$~sec, including $0.11$~sec to compute the disparity and $0.03$~sec for refocusing.
Tab.~\ref{tab:timing} shows how adaptive downsampling helps to reduce runtime complexity in particular for wider aperture values in contrast to naive refocusing, where the runtime grows exponentially with respect to the aperture size.

\vspace{-1.0em}
\paragraph{Limitations.}\noindent
The current approach has some limitations. The first one is inherent to all approaches relying on pyramidal depth estimation: small details that are lost at the lowest scale are difficult to recover at the upper scale which is why thin structures are problematic for our depth estimation (cf. the mirror of the truck in Fig.~\ref{fig:kitti}). The second observation we make is that StereoNet does not perform as well on real data as on synthetic data. Apart from the obvious difficulties (e.g., specularities, rectification errors, noise, optical aberrations) real data embeds, we also believe the sparse ground truth provided by the projected \textit{Lidar} data used for supervision does not encourage the network to refine well at the object boundaries.
Finally, the refocusing part suffers from the same problems as all image-based shallow DoF rendering techniques: it does not handle a defocus foreground very well. This is due to the fact that we miss occluded information when blurring from one view only.

\begin{table}[t]
    \centering
    \begin{tabular}{L{1.9cm}C{1.4cm}C{1.4cm}C{1.4cm}C{1.4cm}}
        \textbf{FPS} for $a=$     & 0.1 & 0.2  & 0.5 & 0.8 \\
        \toprule
        $k_{max}=\infty$     & \bf{76} & 17 & 2 & 0.4 \\
        $k_{max}=11$ & \bf{76} & \bf{38} & \bf{23} & \bf{18}\ \\
        \bottomrule
        %0.1 & 0.013145s & 0.013223s\\
        %0.2 & 0.059643s & 0.026297s\\
        %0.5 & 0.621335s & 0.043122s\\
        %0.8 & 2.334184s & 0.055684s\\
    \end{tabular}
    %\vspace{-1.0em}
    \caption{Runtime evaluation of the layered refocusing pipeline without ($k_{max}=\infty$) and with ($k_{max}=11$) adaptive downsampling on images of~\cite{geiger2012we}. Frames per second are evaluated by average of 10 runs.}
    \label{tab:timing}
\end{table}

\section{Conclusion}\noindent
The entanglement of stereo-based depth estimation and refocusing proves to be a promising solution for the task of efficient scene-aware image reblurring with appealing bokeh.
Future improvements can address a fusion with segmentation methods to enhance boundary precision similar to~\cite{nekrasov2018real} who entangle the task of semantic segmentation and depth estimation in real-time.
The lightweight differentiable architecture with the insight about the value given by a sequential approach with depth supervision can be used for a variety of image and video refocusing applications in other vision pipelines that utilize our refocusing module.
For instance in mobile applications, an image pair is taken at one point and different refocusing results may be calculated to help selection by a user afterwards, and the efficiency of our pipeline paves the way to real-time video editing applications on the edge.

\clearpage

\FloatBarrier
{\small
\bibliographystyle{ieee}
\bibliography{literature}
}

%%%%%%%%% BODY TEXT
\section{Additional Results}\noindent
In addition to Figure 6 found in the main paper, Figure~\ref{tab:comparison_img} shows further results on the SceneFlow dataset~\cite{mayer2016large} for the different pipelines, $A_1$, $A_2$, $B$ and $C$, which we propose.

\begin{figure*}[t]
    \centering
    \begin{tabular}{ccccc}
    Ground Truth & Sequential Depth ($A_1$) & Sequential Aperture ($A_2$) & CV Refinement ($B$) & Blur Refinement ($C$)\\

    \includegraphics[width=0.2\textwidth]{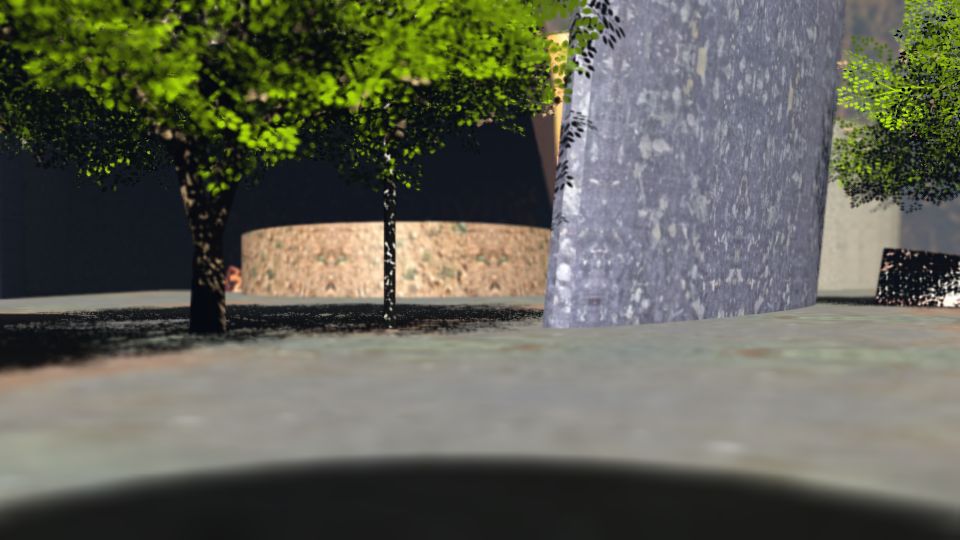}&\includegraphics[width=0.2\textwidth]{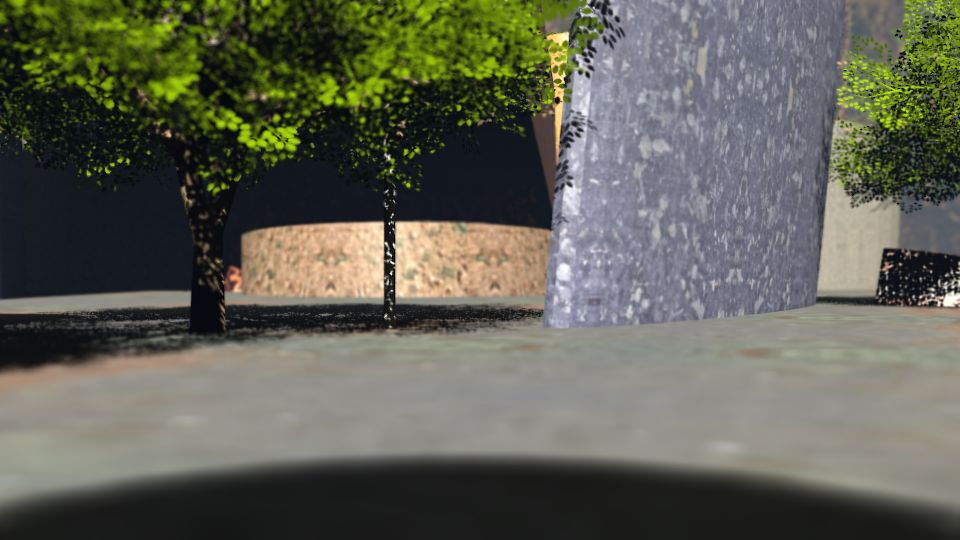}&\includegraphics[width=0.2\textwidth]{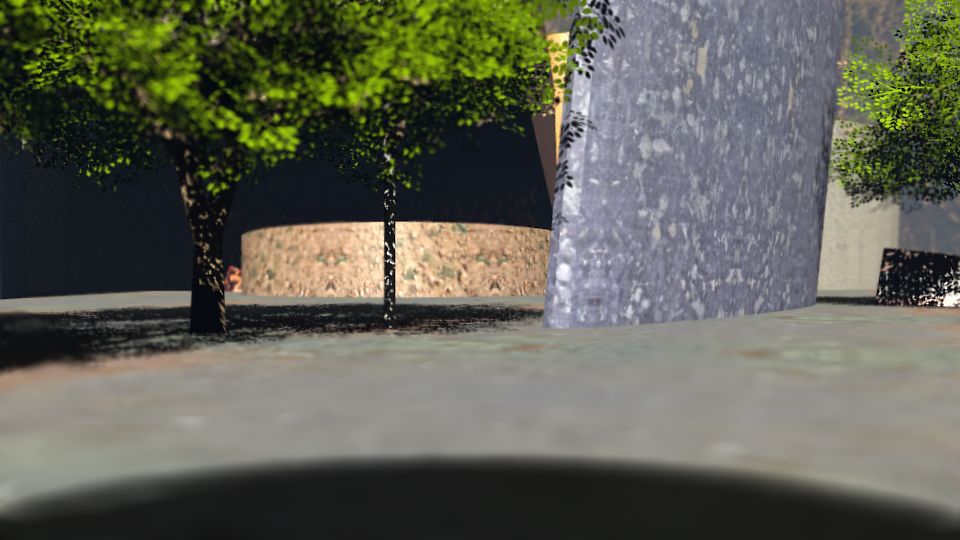}&\includegraphics[width=0.2\textwidth]{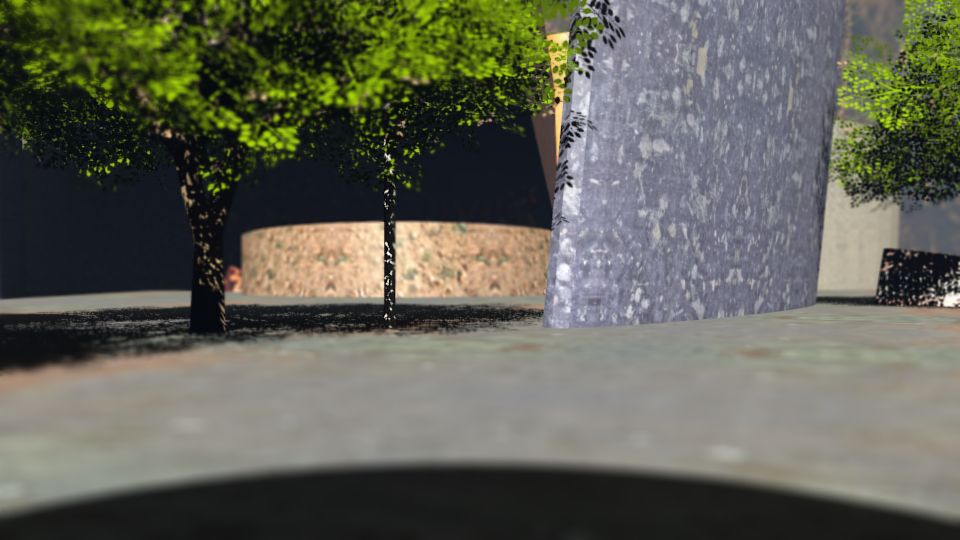}&\includegraphics[width=0.2\textwidth]{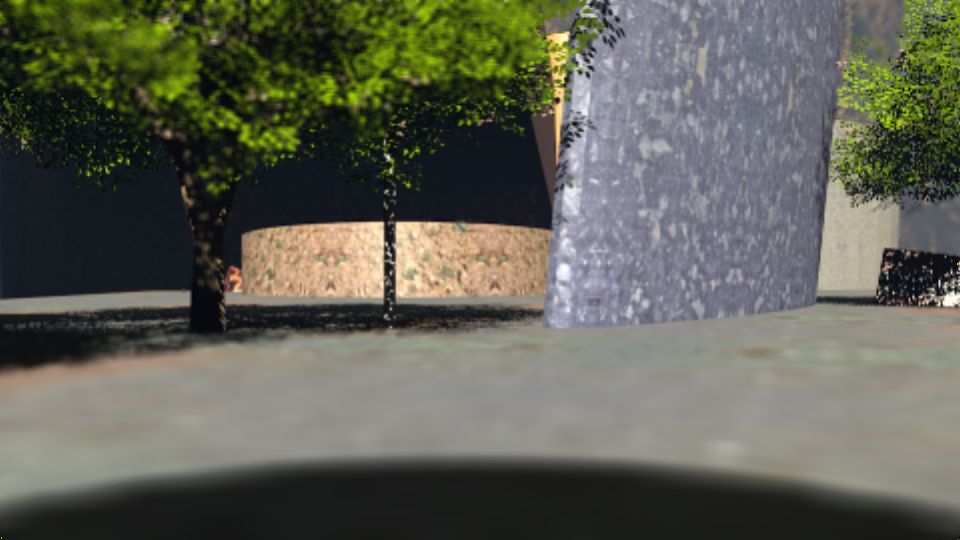}\\
    
    \includegraphics[width=0.2\textwidth]{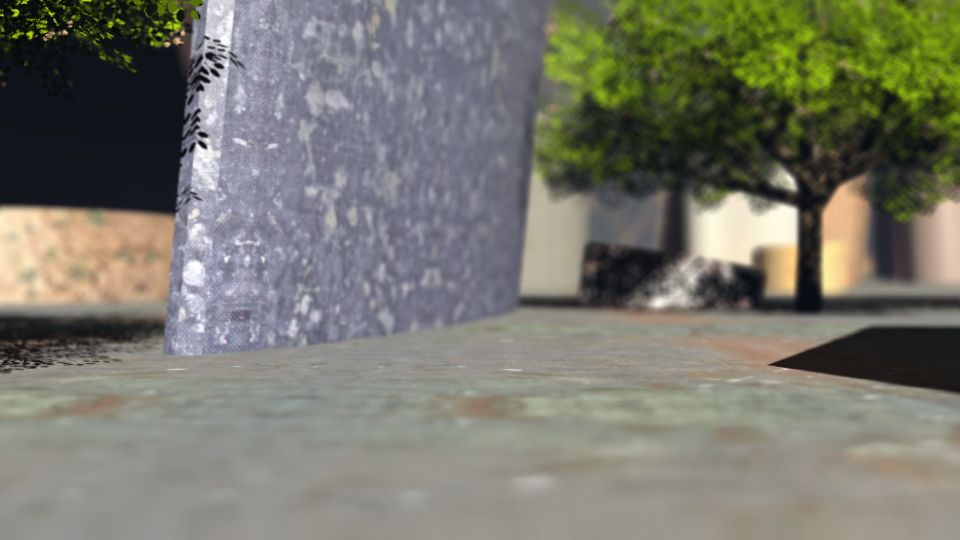}&\includegraphics[width=0.2\textwidth]{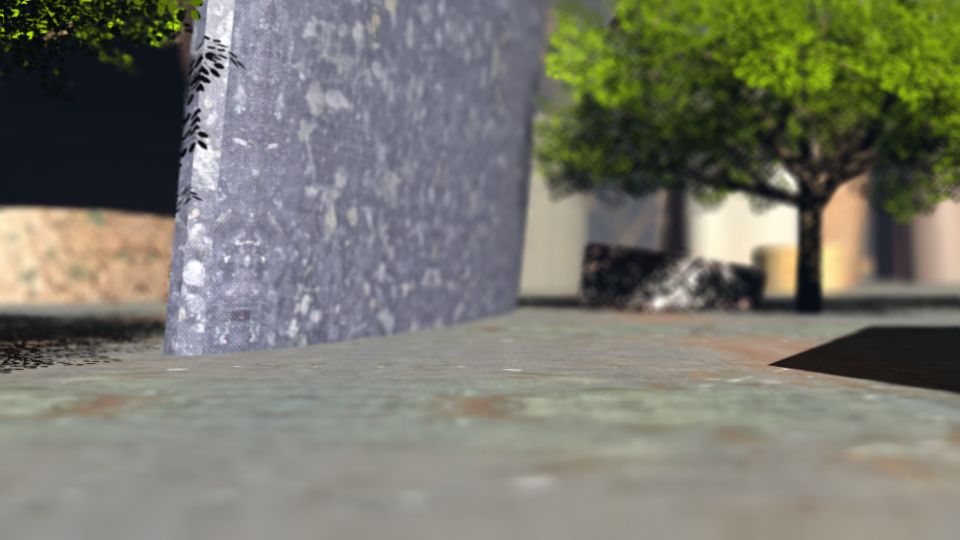}&\includegraphics[width=0.2\textwidth]{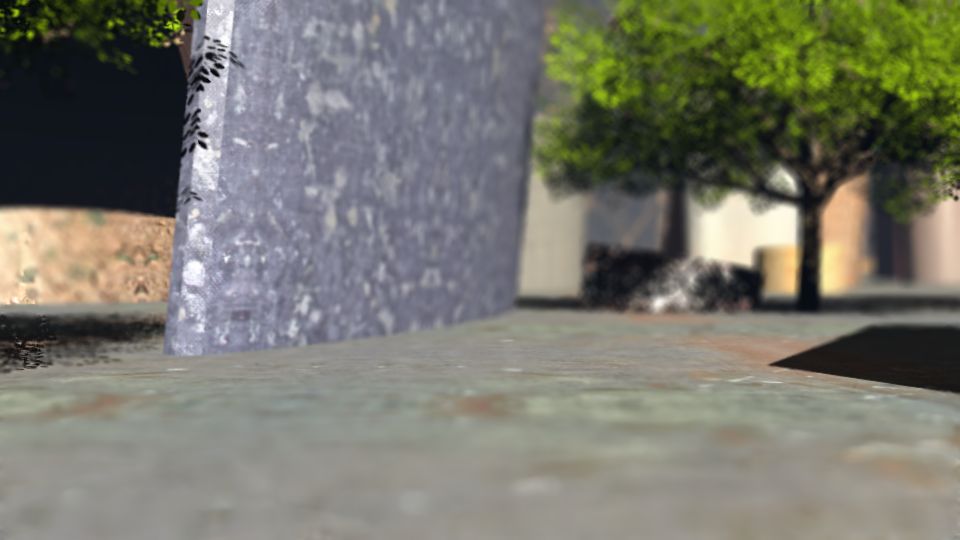}&\includegraphics[width=0.2\textwidth]{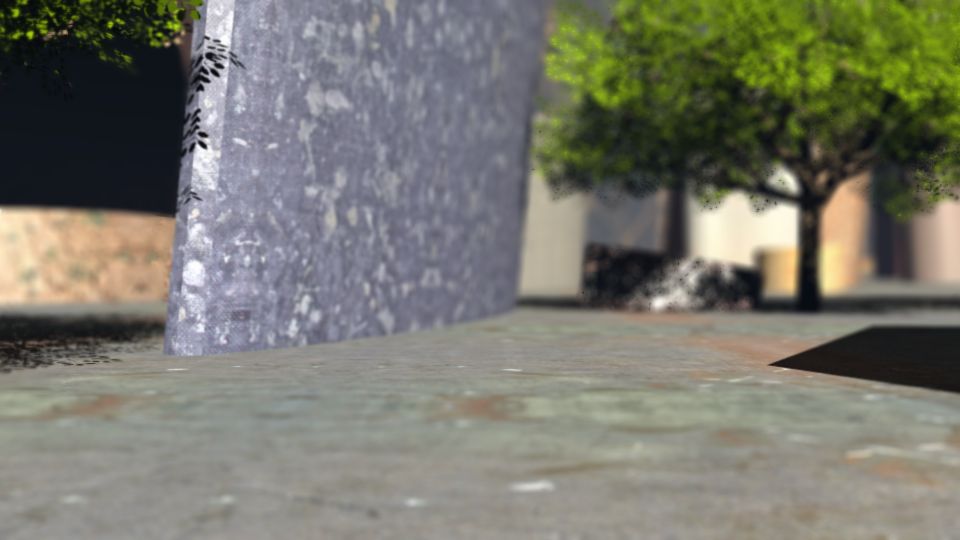}&\includegraphics[width=0.2\textwidth]{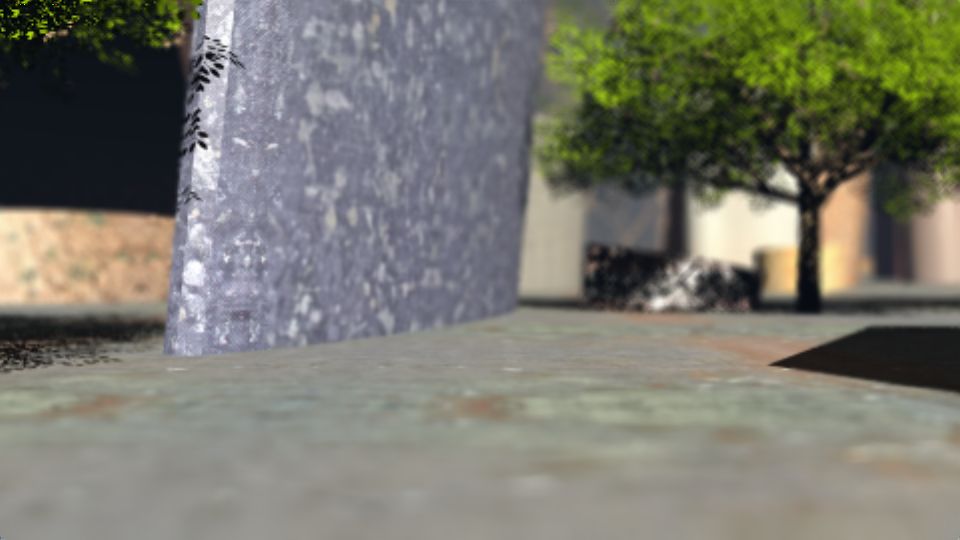}\\

    \includegraphics[width=0.2\textwidth]{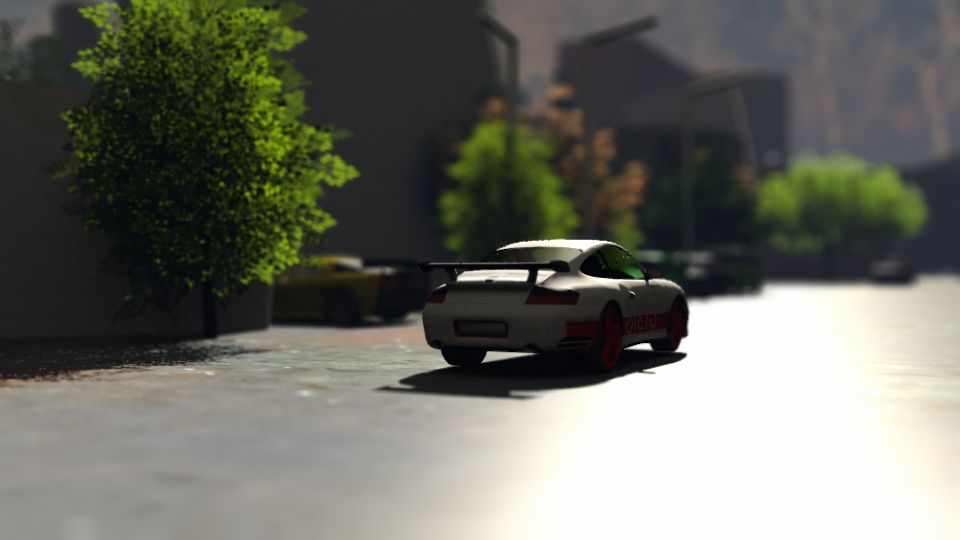}&\includegraphics[width=0.2\textwidth]{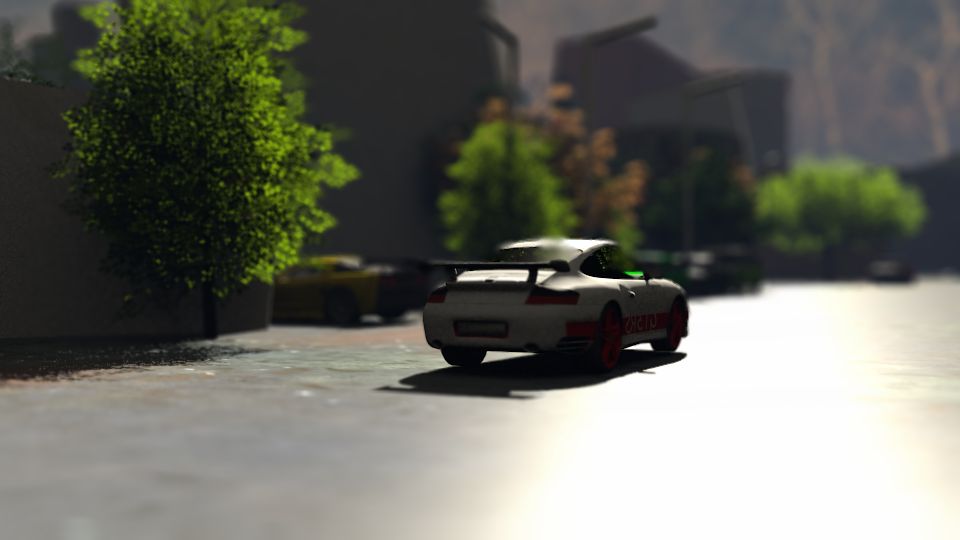}&\includegraphics[width=0.2\textwidth]{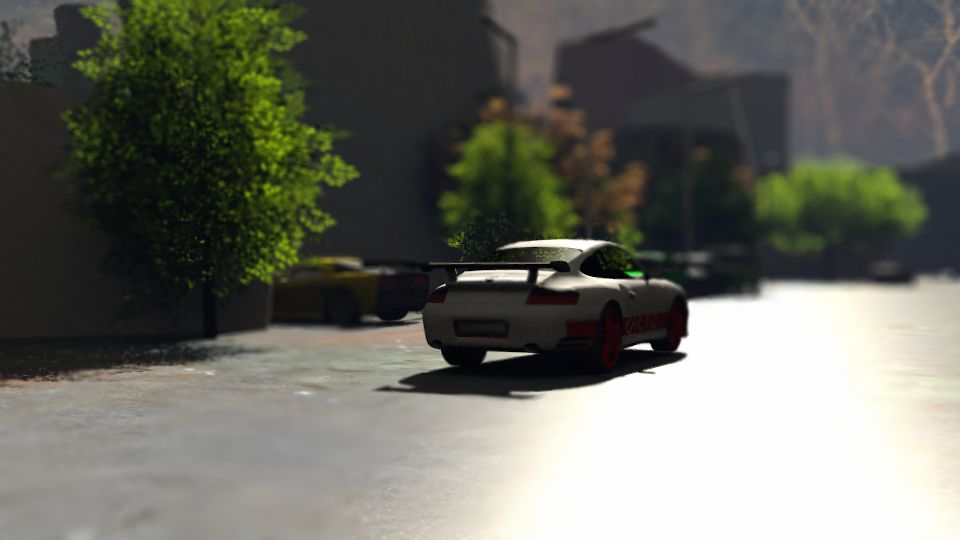}&\includegraphics[width=0.2\textwidth]{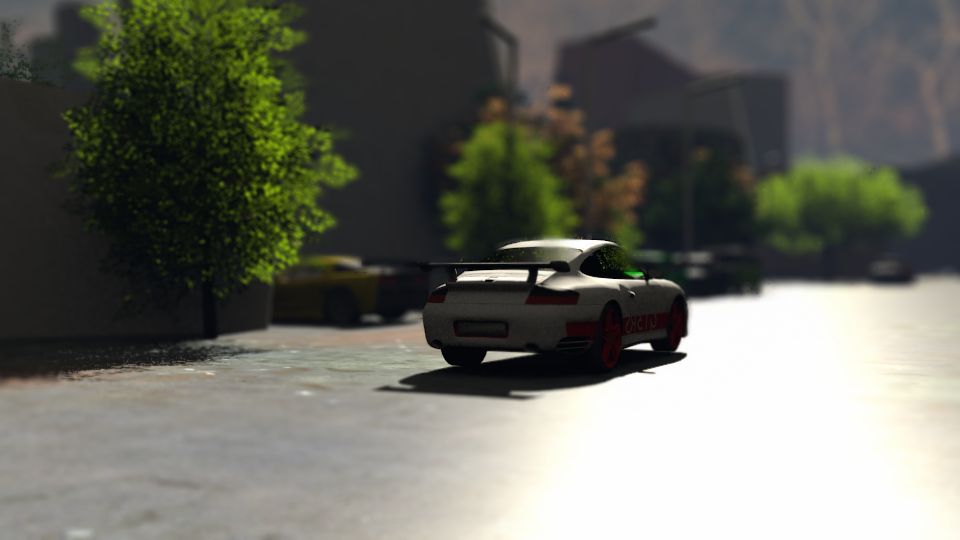}&\includegraphics[width=0.2\textwidth]{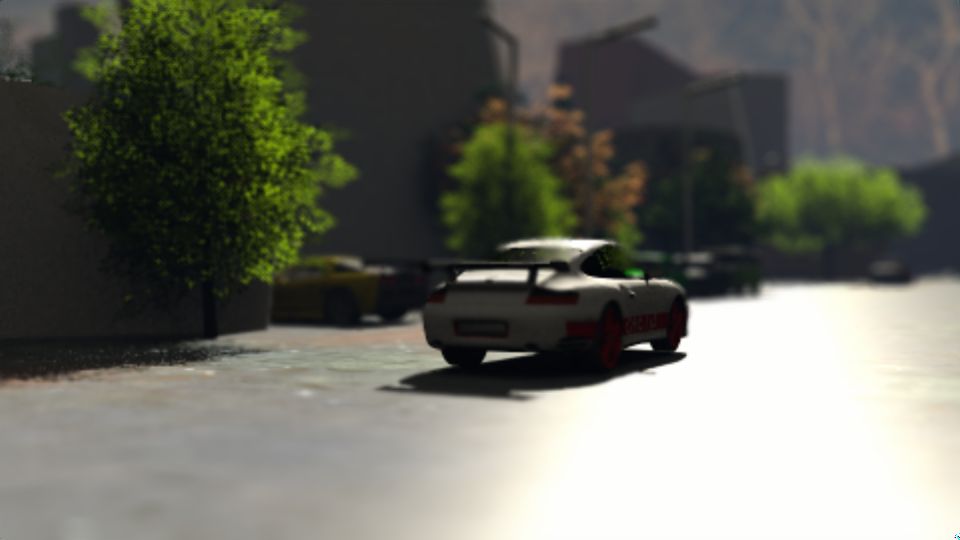}\\
    
    \includegraphics[width=0.2\textwidth]{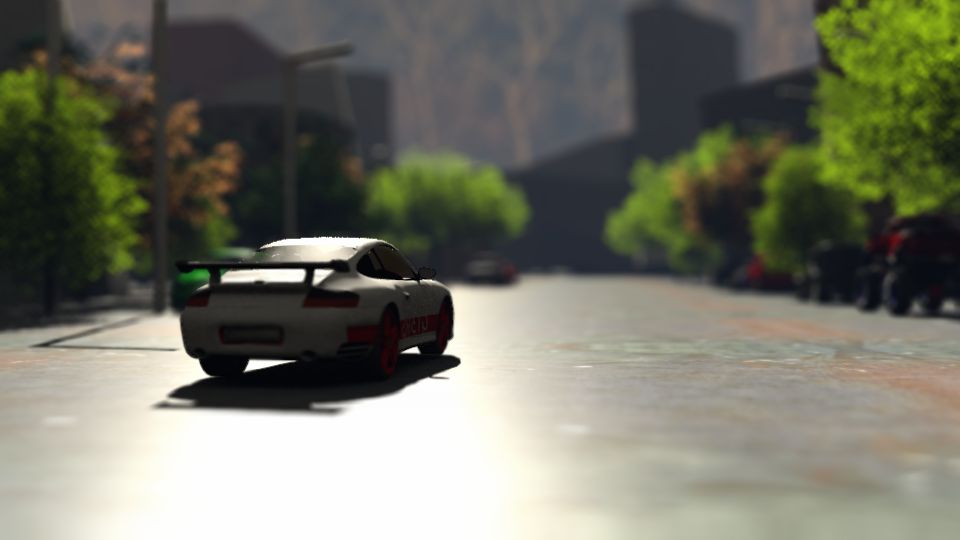}&\includegraphics[width=0.2\textwidth]{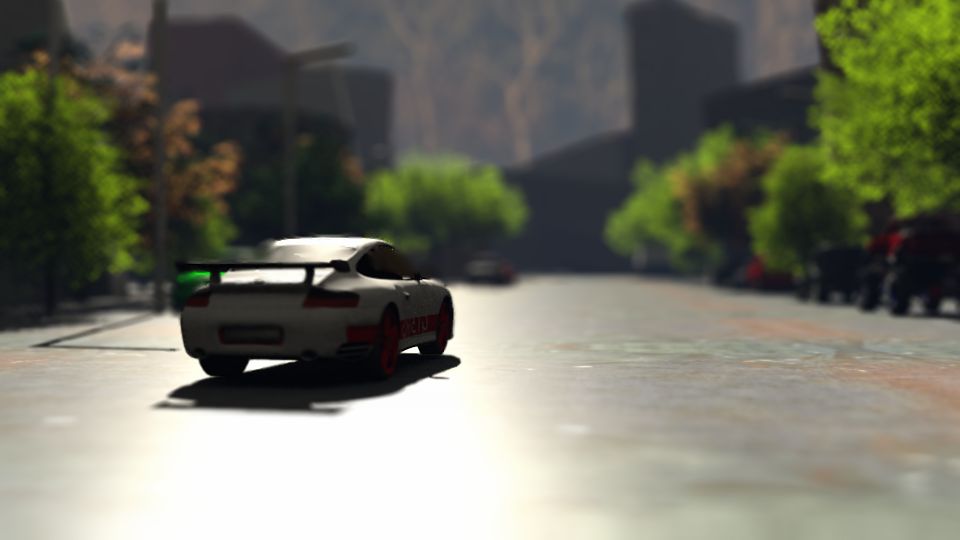}&\includegraphics[width=0.2\textwidth]{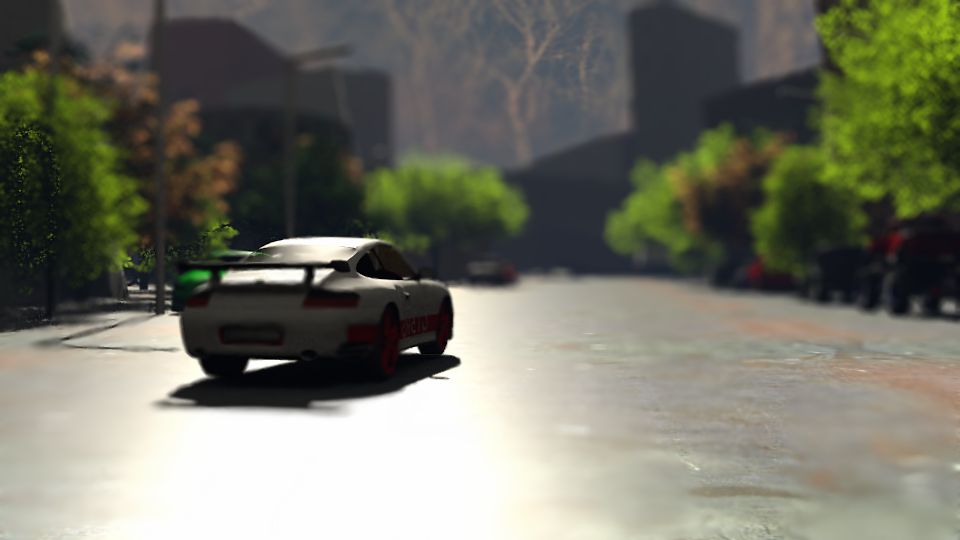}&\includegraphics[width=0.2\textwidth]{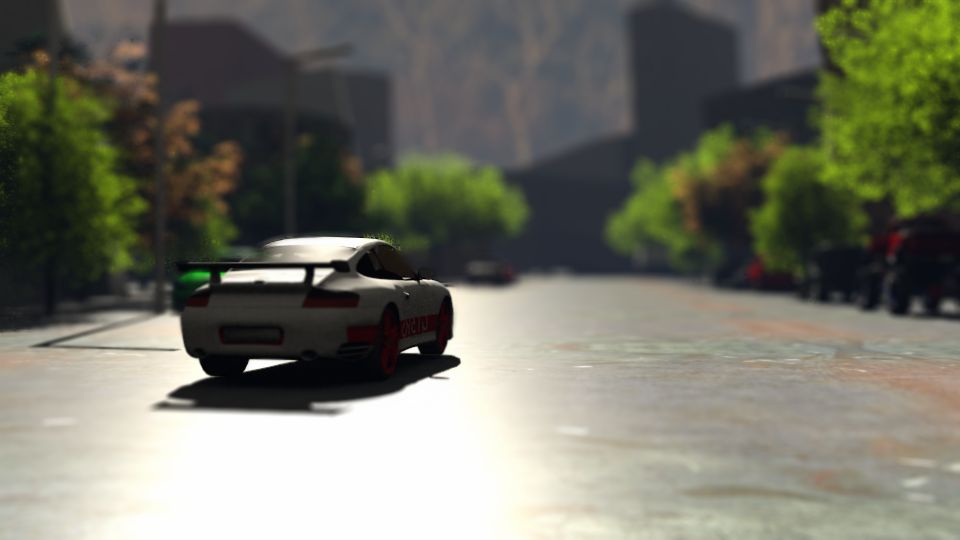}&\includegraphics[width=0.2\textwidth]{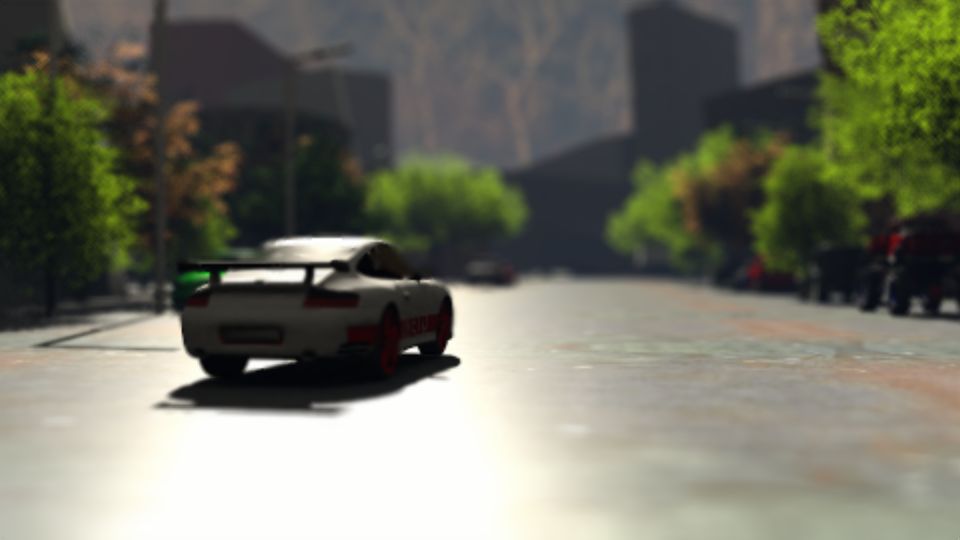}\\
    
    \includegraphics[width=0.2\textwidth]{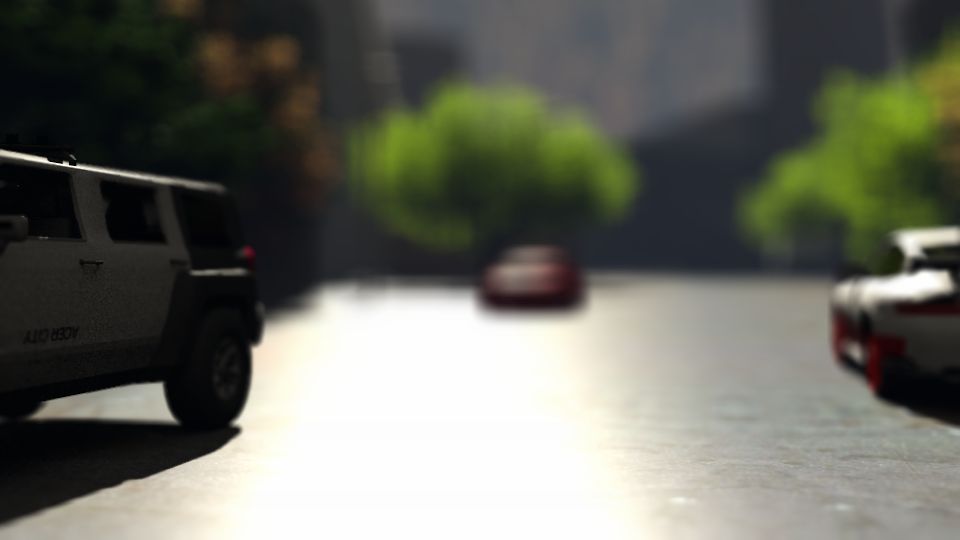}&\includegraphics[width=0.2\textwidth]{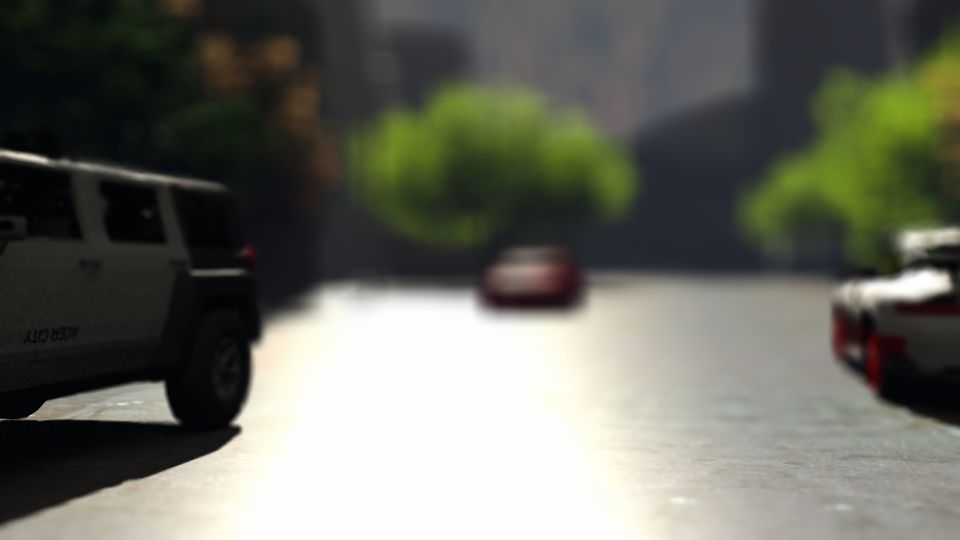}&\includegraphics[width=0.2\textwidth]{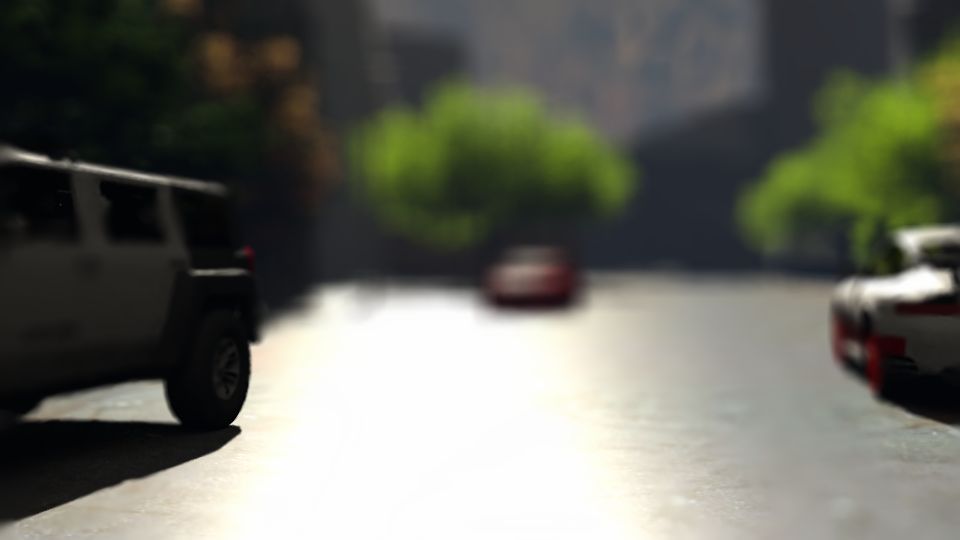}&\includegraphics[width=0.2\textwidth]{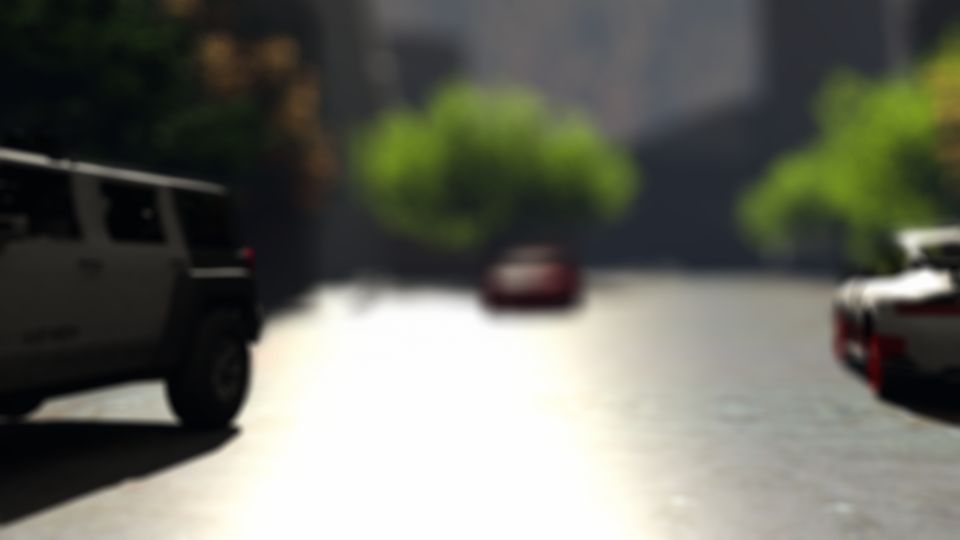}&\includegraphics[width=0.2\textwidth]{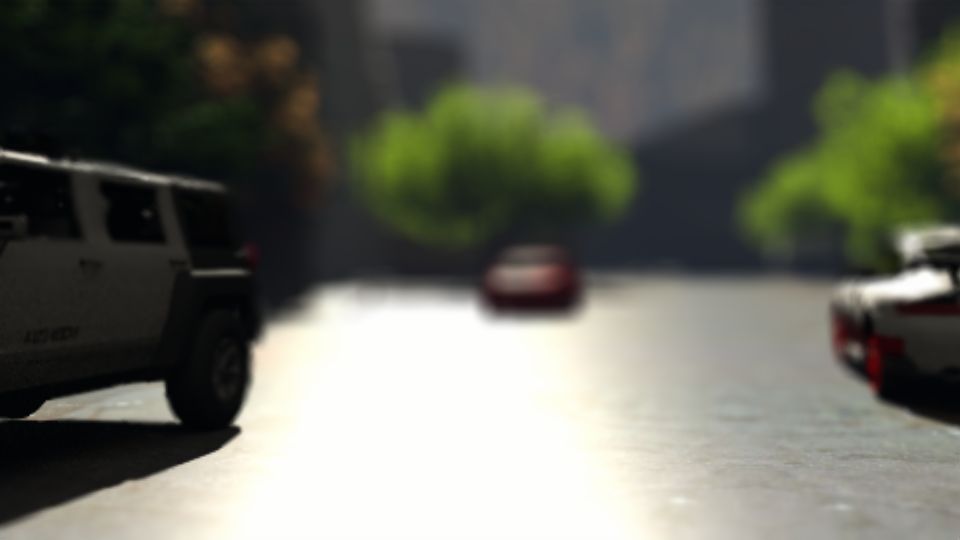}\\
    
    \includegraphics[width=0.2\textwidth]{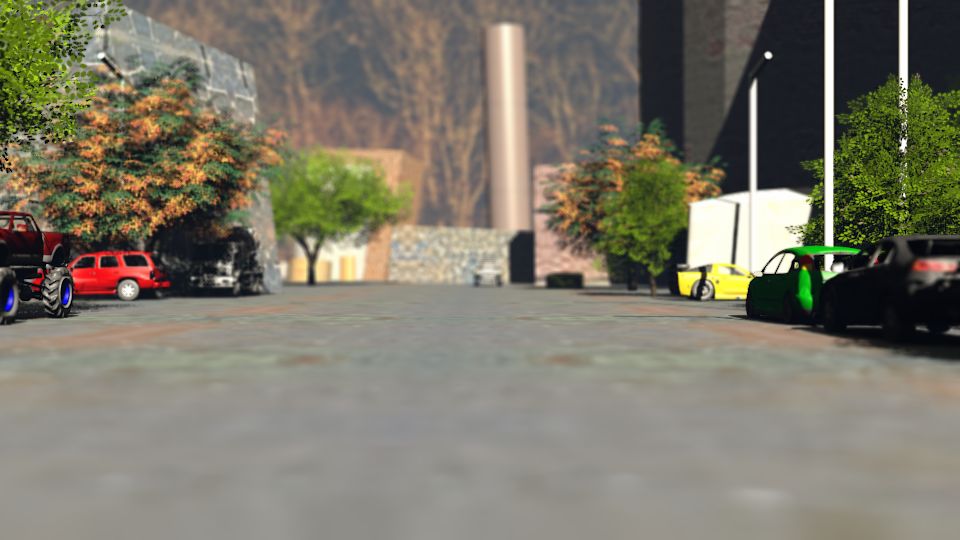}&\includegraphics[width=0.2\textwidth]{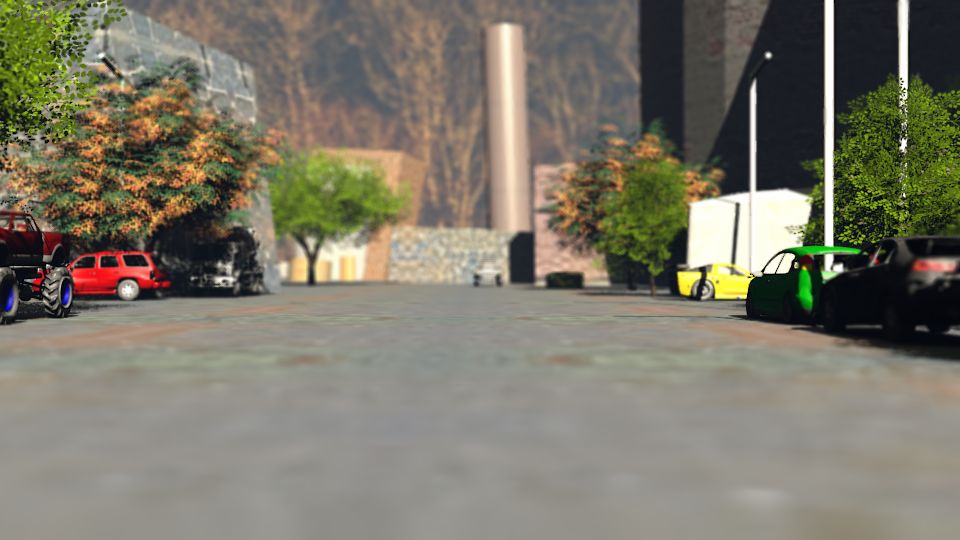}&\includegraphics[width=0.2\textwidth]{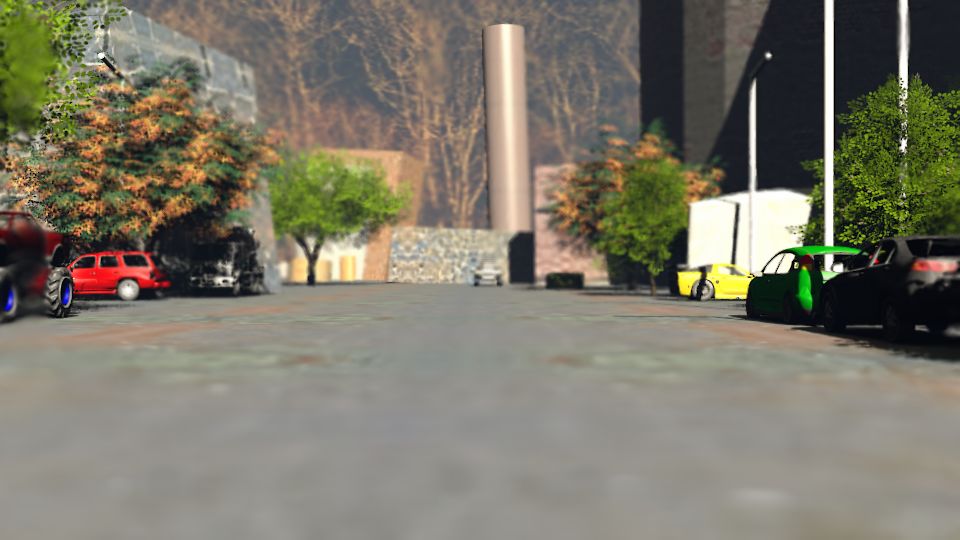}&\includegraphics[width=0.2\textwidth]{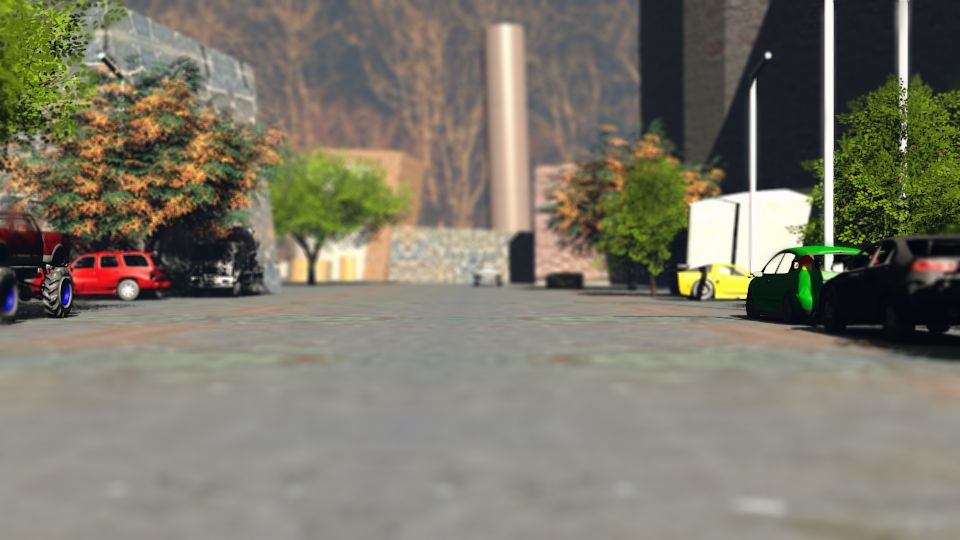}&\includegraphics[width=0.2\textwidth]{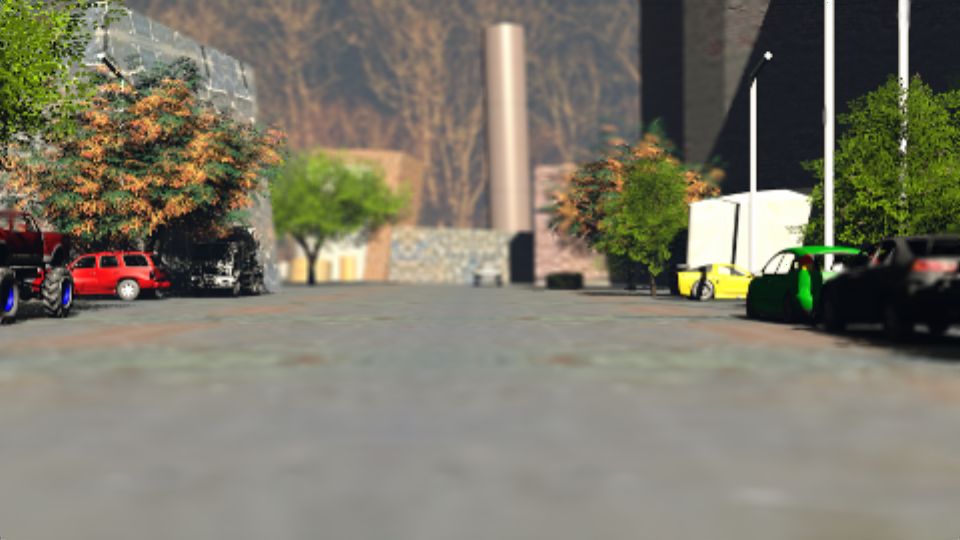}\\
    
    \includegraphics[width=0.2\textwidth]{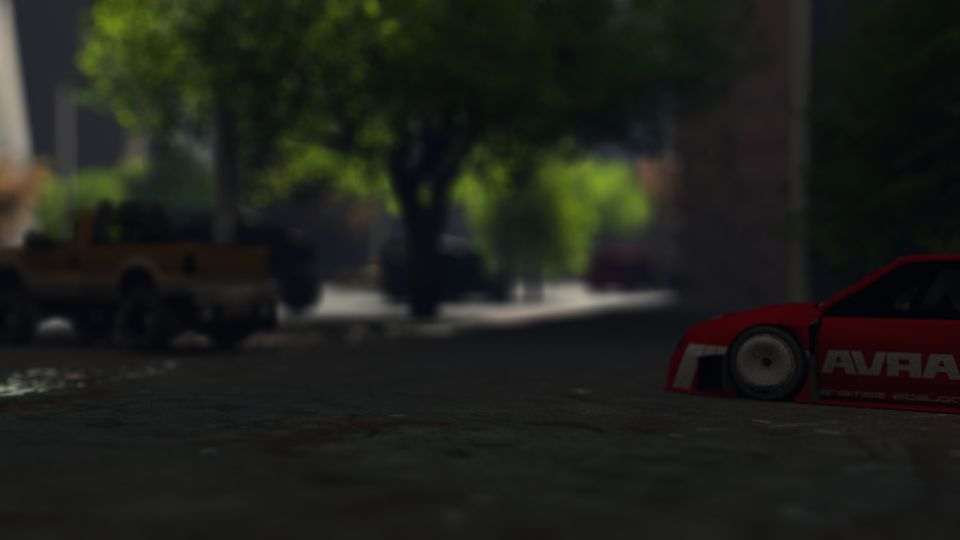}&\includegraphics[width=0.2\textwidth]{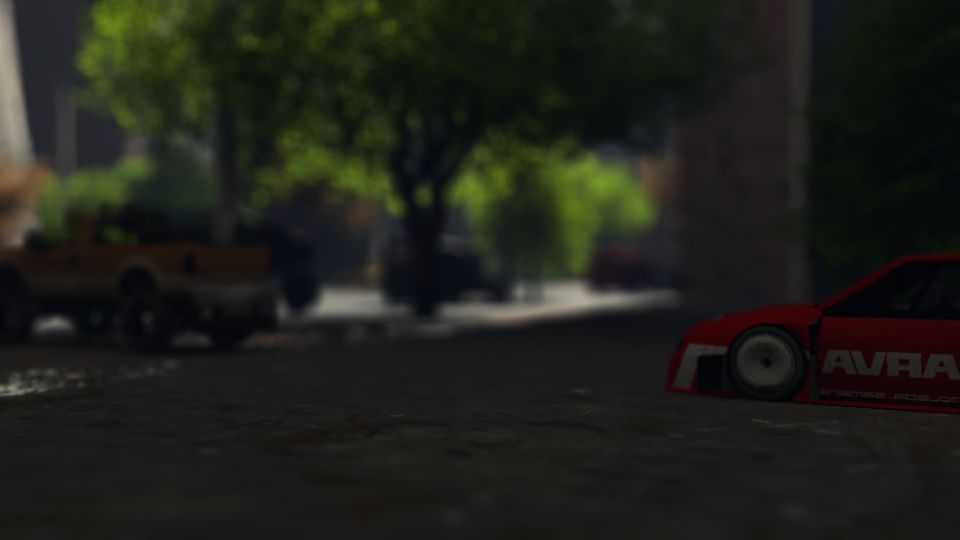}&\includegraphics[width=0.2\textwidth]{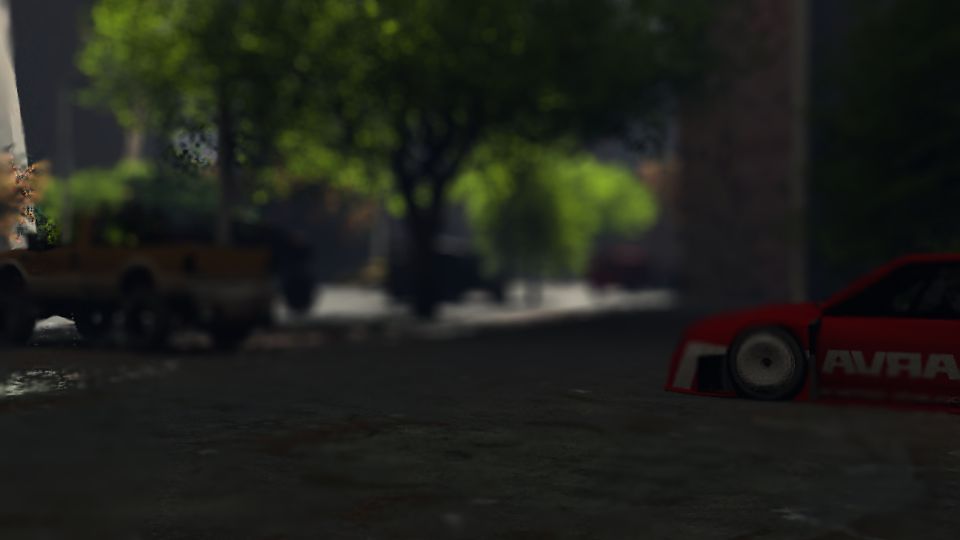}&\includegraphics[width=0.2\textwidth]{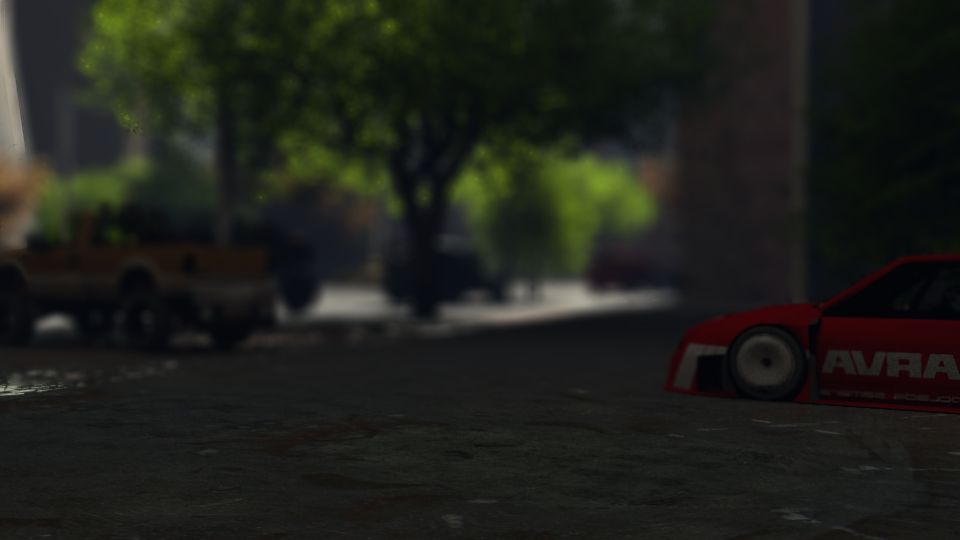}&\includegraphics[width=0.2\textwidth]{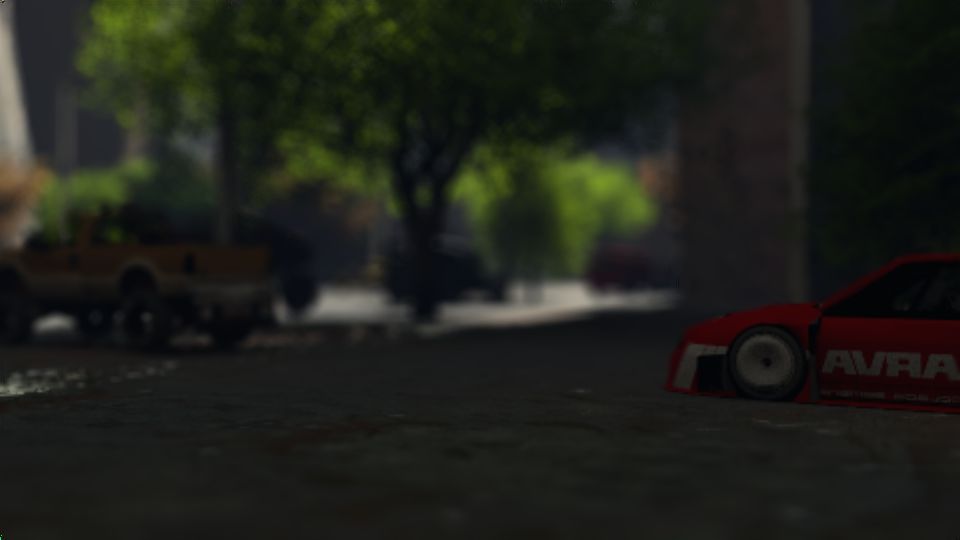}\\
    
    \includegraphics[width=0.2\textwidth]{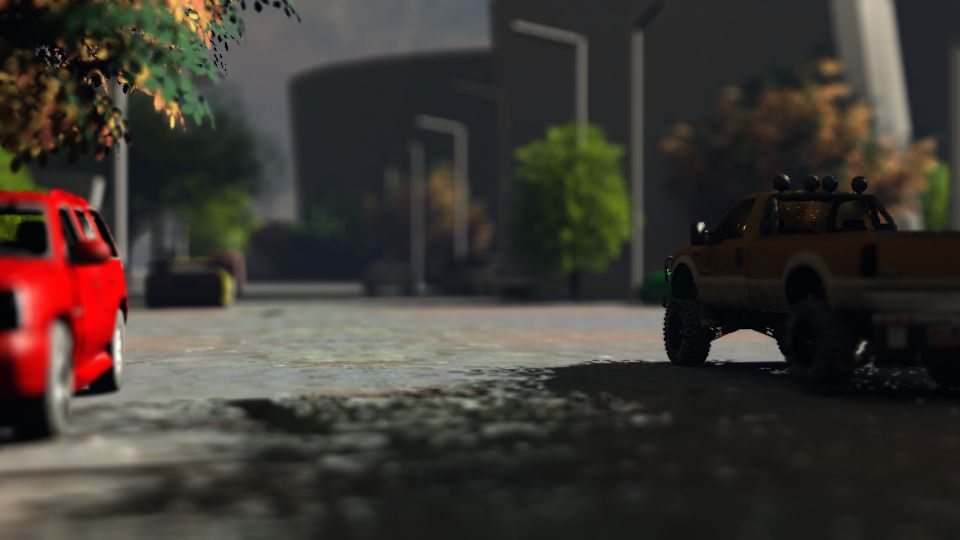}&\includegraphics[width=0.2\textwidth]{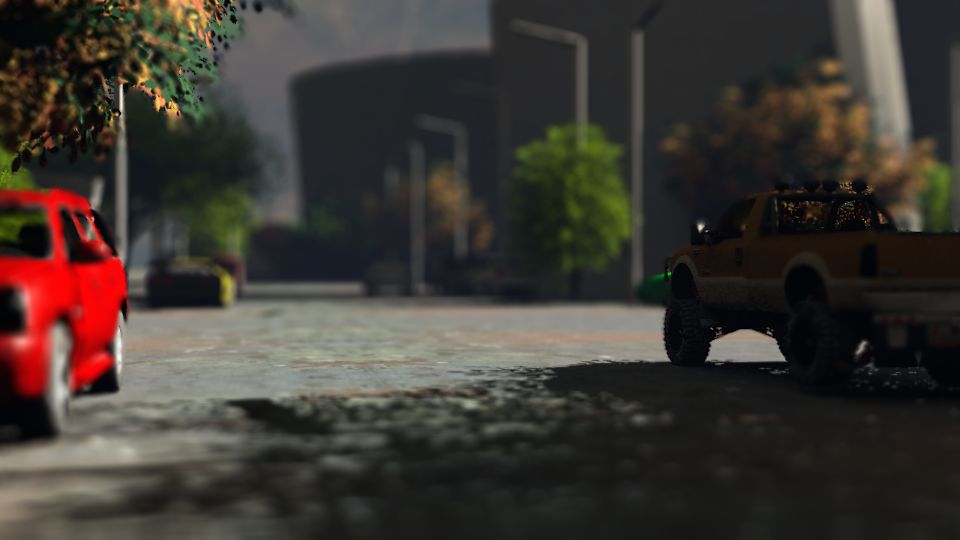}&\includegraphics[width=0.2\textwidth]{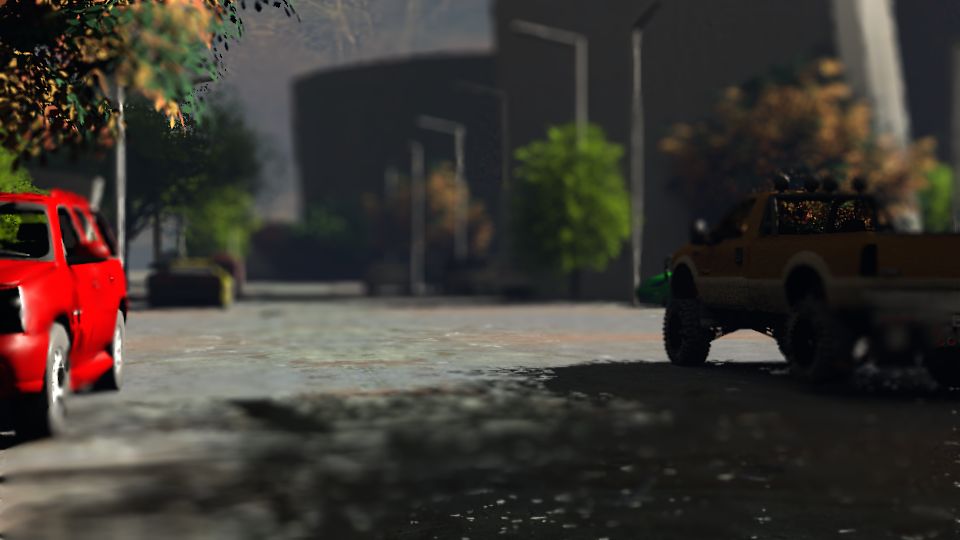}&\includegraphics[width=0.2\textwidth]{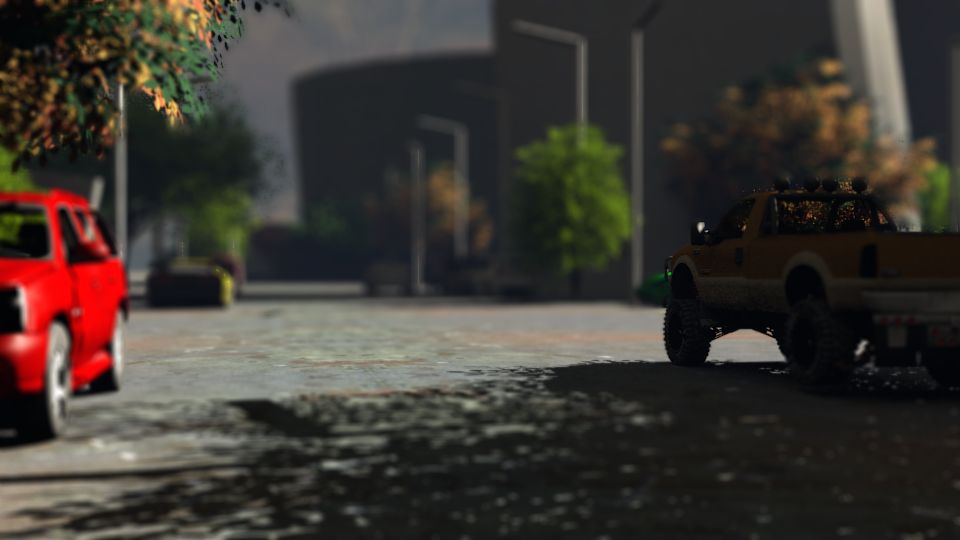}&\includegraphics[width=0.2\textwidth]{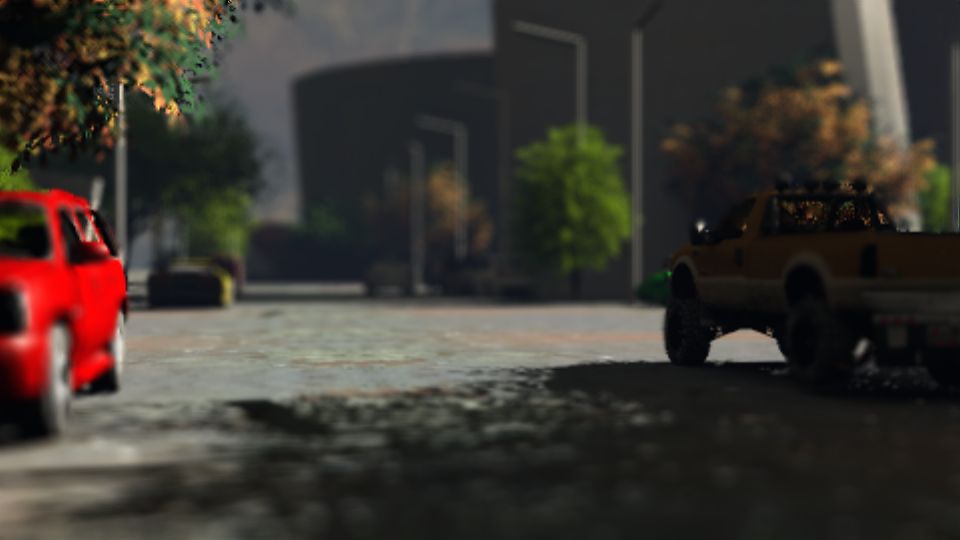}\\

    \end{tabular}
    %\vspace{-1.0em}
    \caption{Comparison of the four different approaches for the rest of the test set. Each column corresponds to one method and each row to one test image. We display on the very left column the ground truth image refocused using the provided ground truth disparity. We invite the reader to zoom-in to see the details.}
    % \caption{Comparison of our four different approaches using the remaining test set images, not shown in the main paper. Each column corresponds to one method and each row to one test image. The leftmost column shows the ground truth image, refocused using the provided ground truth disparity. We invite the reader to zoom-in to see the details.}
    \label{tab:comparison_img_supp}
\end{figure*}

\end{document}